\documentclass[runningheads]{llncs}
\usepackage{graphicx}
\usepackage{amsmath,amssymb} 
\usepackage{color}
\usepackage[width=122mm,left=12mm,paperwidth=146mm,height=193mm,top=12mm,paperheight=217mm]{geometry}
\usepackage{mathrsfs}
\usepackage{afterpage}
\usepackage{float}
\usepackage{booktabs}  
\usepackage[colorlinks, linkcolor=blue, anchorcolor=blue,  citecolor=blue]{hyperref}
\begin{document}

\newcommand{\etal}{et al.}

\newcommand{\Yijun}[1]{\textcolor{black}{#1}}

\newcommand{\jimei}[1]{\textcolor{red}{[jimei: {#1}]}}
\newcommand{\Chen}[1]{\textcolor{red}{[Chen: {#1}]}}

\pagestyle{headings}
\mainmatter

\title{Flow-Grounded Spatial-Temporal Video Prediction from Still Images} 

\titlerunning{Flow-Grounded Spatial-Temporal Video Prediction from Still Images}

\authorrunning{Y. Li, C. Fang, J. Yang, Z. Wang, X. Lu, M.-H. Yang}

\author{Yijun Li$^1$, Chen Fang$^2$, Jimei Yang$^2$, Zhaowen Wang$^2$\\ Xin Lu$^2$, Ming-Hsuan Yang$^{1,3}$}


\institute{$^1$University of California, Merced~~~~~$^2$Adobe Research~~~~~$^3$Google\\
	\email{\{yli62,mhyang\}@ucmerced.edu}~~~~\email{\{cfang,jimyang,zhawang,xinl\}@adobe.com}
}

\maketitle

\begin{abstract}
Existing video prediction methods mainly rely on observing multiple historical frames or focus on predicting the next one-frame. In this work, we study the problem of generating consecutive multiple future frames by observing one single still image only. We formulate the multi-frame prediction task as a multiple time step flow (multi-flow) prediction phase followed by a flow-to-frame synthesis phase. The multi-flow prediction is modeled in a variational probabilistic manner with spatial-temporal relationships learned through 3D convolutions. The flow-to-frame synthesis is modeled as a generative process in order to keep the predicted results lying closer to the manifold shape of real video sequence. Such a two-phase design prevents the model from directly looking at the high-dimensional pixel space of the frame sequence and is demonstrated to be more effective in predicting better and diverse results. 
Extensive experimental results on videos with different types of motion show that the proposed algorithm performs favorably against existing methods in terms of quality, diversity and human perceptual evaluation.

\end{abstract}

\section{Introduction}

\begin{figure}[t]
\centering
\begin{tabular}{c@{\hspace{0.005\linewidth}}c@{\hspace{0.005\linewidth}}c@{\hspace{0.005\linewidth}}c@{\hspace{0.005\linewidth}}c@{\hspace{0.005\linewidth}}c@{\hspace{0.005\linewidth}}c@{\hspace{0.005\linewidth}}c@{\hspace{0.005\linewidth}}c@{\hspace{0.005\linewidth}}c}

    \includegraphics[width = .103\linewidth]{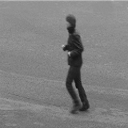} &
    
    \hspace{1pt}\vrule\hspace{1pt}
    
    \includegraphics[width = .103\linewidth]{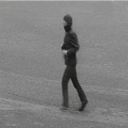} &
    \includegraphics[width = .103\linewidth]{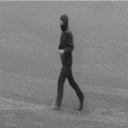} &
    \includegraphics[width = .103\linewidth]{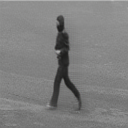} &
    \includegraphics[width = .103\linewidth]{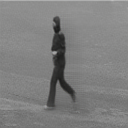} &
    \includegraphics[width = .103\linewidth]{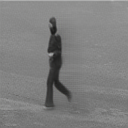} &
    \includegraphics[width = .103\linewidth]{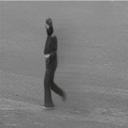} &
    \includegraphics[width = .103\linewidth]{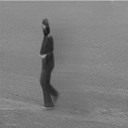} &
    \includegraphics[width = .103\linewidth]{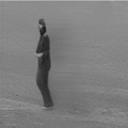} & \\

    \includegraphics[width = .103\linewidth]{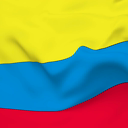} &
    
    \hspace{1pt}\vrule\hspace{1pt}
    
    \includegraphics[width = .103\linewidth]{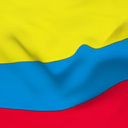} &
    \includegraphics[width = .103\linewidth]{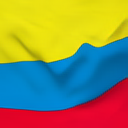} &
    \includegraphics[width = .103\linewidth]{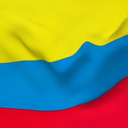} &
    \includegraphics[width = .103\linewidth]{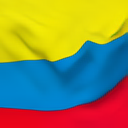} &
    \includegraphics[width = .103\linewidth]{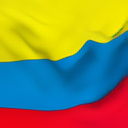} &
    \includegraphics[width = .103\linewidth]{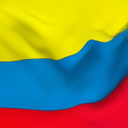} &
    \includegraphics[width = .103\linewidth]{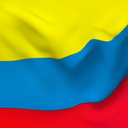} &
    \includegraphics[width = .103\linewidth]{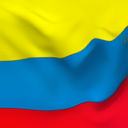} & \\

    { t=0  }& { t=1  }& { t=2  }& { t=3  }& { t=4  }& { t=5  }& { t=6  }& { t=7  }& { t=8  } \\
\end{tabular}
\caption{Multi-step future sequences generated by our algorithm (t=1$\sim$8) conditioned on one single still image (t=0). Images are of size 128$\times$128. 
}
\label{fig:teaser}
\end{figure}

Part of our visual world constantly experiences situations that require us to forecast what will happen over time by observing one still image from a single moment.
Studies in neuroscience show that this \emph{preplay} activity might constitute an automatic prediction mechanism in human visual cortex~\cite{ekman2017time}.
Given the great progress in artificial intelligence, researchers also begin to let machines learn to perform such a predictive activity for various applications. 
For example in Figure~\ref{fig:teaser}(top), from a snapshot by the surveillance camera, the system is expected to predict the man's next action which could be used for safety precautions. 
Another application in computational photography is turning still images into vivid cinemagraphs for aesthetic effects, as shown in Figure~\ref{fig:teaser}(bottom).

In this work, we mainly study how to generate pixel-level future frames in multiple time steps given one still image.
A number of existing prediction models~\cite{mathieu-ICLR-2016,convLSTM-NIPS-2015,mcnet-ICLR-2017,drnet-NIPS-2017} are under the assumption of observing a short video sequence ($>$1 frame). Since multiple historical frames explicitly exhibit obvious motion cues, most of them use deterministic models to render a fixed future sequence.
In contrast, our single-image based prediction task, without any motion information provided, implies that there are obvious uncertainties existed in both spatial and temporal domains.
Therefore we propose a probabilistic model based on a conditional variational autoencoder (cVAE) to model the uncertainty. Our probabilistic model has two unique features. First, it is a 3D-cVAE model, i.e., the autoencoder is designed in a spatial-temporal architecture with 3D convolution layers. 
The 3D convolutional layer~\cite{c3d-ICCV-2015}, which takes a volume as input, is able to capture correlations between the spatial and temporal dimension of signals, thereby rendering distinctive spatial-temporal features for better predictions.
Second, the output of our model is optical flows which characterize the spatial layout of how pixels are going to move step by step.
Different from other methods that predict trajectories~\cite{walker2016uncertain}, frame differences~\cite{crossconv-NIPS-2016} or frame pixels~\cite{drnet-NIPS-2017}, the flow is a more natural and general representation of motions. It serves as a relatively low-dimensional reflection of high-level structures and can be obtained in an unsupervised manner.


With the predicted flows, we next formulate the full frame synthesis as a generation problem. 
Due to the existence of occlusions, flow-based pixel-copying operations (e.g., warping) are obviously ineffective here.
The model should be capable of ``imagining'' the appearance of future frames and removing the unnecessary parts in the previous frame at the same time.
Therefore we propose a generative model \emph{Flow2rgb} to generate pixel-level future frames.
Such a model is non-trivial and is demonstrated to be effective in keeping the generated sequence staying close to the manifold of real sequences (Figure~\ref{fig:embedding_manifold}). 
Overall, we formulate the multi-frame prediction task as a multiple time step flow prediction phase followed by a flow-to-frame generation phase. Such a two-phase design prevents the model from directly looking at the high-dimensional pixel space of the frame sequence and is demonstrated to be more effective in predicting better results.
During the testing, by drawing different samples from the learned latent distribution, our approach can also predict diverse future sequences.

%

The main contributions of this work are summarized as follows:
\begin{itemize}
\item We propose a spatial-temporal conditional VAE model (3D-cVAE) to predict future flows in multiple time steps. The diversity in predictions is realized by drawing different samples from the learned distribution.

\item We present a generative model that learns to generate the pixel-level appearance of future frames based on predicted flows. 

\item We demonstrate the effectiveness of our method for predicting sequences that contain both articulated (e.g., humans) objects and dynamic textures (e.g., clouds).
\end{itemize}

\section{Related Work}

\paragraph{\bf Action prediction.}
The macroscopic analysis of prediction based on the given frame(s) can be predicting what event is going to happen~\cite{yuen2010data,lan2014hierarchical,hoai2014max}, trajectory paths~\cite{kitani2012activity}, or recognizing the type of human activities~\cite{vondrick2016anticipating,walker-ICCV-2015}. 
Some of early methods are supervised, requiring labels (e.g., bounding boxes) of the moving object. Later approaches~\cite{walker-ICCV-2015} realize the unsupervised way of prediction by relying on the context of scenes.  
However, these approaches usually only provide coarse predictions of how the future will evolve and are unable to tell richer information except for a action (or event) label. 

\paragraph{\bf Pixel-level frame prediction.}~Recent prediction methods move to the microcosmic analysis of more detailed information in the future. 
This is directly reflected by requiring the pixel-level generation of future frames in multiple time steps.
With the development of deep neural networks, especially when recursive modules are extensively used, predicting realistic future frames has being dominated.
Much progress has been made in the generated quality of future outputs by designing different network structures~\cite{srivastava-ICML-2015,oh-NIPS-2015action,mathieu-ICLR-2016,babaeizadeh2017stochastic,finn2017deep} or using different learning techniques, including adversarial loss~\cite{videoGAN-NIPS-2016,liang-ICCV-2017}, motion/content separation~\cite{mcnet-ICLR-2017,mocoGAN-2017,drnet-NIPS-2017}, and transformation parameters~\cite{finn2016unsupervised,carl-CVPR-2017transformer}.

Our work also aims at accurate frame predictions but the specific setting is to model the uncertainties of multi-frame prediction given a single still image as input.
In terms of multi-frame predictions conditioning on still images, closest work to ours are~\cite{johnny-CVPR-2017,ruben-ICML-2017}. However,~\cite{johnny-CVPR-2017} only predicts the pose information and the proposed model is deterministic. The work in \cite{ruben-ICML-2017} also estimates poses first and then use an image-analogy strategy to generate frames. But their pose generation step relies on observing multiple frames.
%
%
\Yijun{
%
%
Moreover, both approaches employ the recursive module (e.g., recurrent neural networks) for consecutive predictions which may overemphasize on learning the temporal information only.
Instead, we use the 3D convolutional layer~\cite{c3d-ICCV-2015} which takes a volume as input. Since both spatial and temporal information are encoded together, the 3D convolution can generally capture correlations between the spatial and temporal dimension of signals, thereby rendering distinctive spatial-temporal features~\cite{c3d-ICCV-2015}.
}
In addition, both~\cite{johnny-CVPR-2017,ruben-ICML-2017} focus on human dynamics while our work targets on both articulated objects and dynamic textures.

In terms of modeling future uncertainties, two methods~\cite{crossconv-NIPS-2016,walker2016uncertain} are closely related. 
However, Xue \etal~\cite{crossconv-NIPS-2016} only model the uncertainty in the next one-step prediction.
If we iteratively run the one-step prediction model for multi-step predictions, the frame quality will degrade fast through error accumulations, due to the lack of temporal relationships modeling between frames.
Though Walker \etal~\cite{walker2016uncertain} could keep forecasting over the course of one second, instead of predicting real future frames, it only predicts the dense trajectory of pixels.
Also such a trajectory-supervised modeling requires laborious human labeling.
Different from these methods, our approach integrates the multi-frame prediction and uncertainty modeling in one model.

\paragraph{\bf Dynamic textures.}~The above-mentioned methods mainly focus on the movement of articulated objects (e.g., human). In contrast, dynamic textures often exhibit more randomness in the movement of texture elements.
Both traditional methods based on linear dynamical systems~\cite{doretto-IJCV-2003,yuan2004synthesizing} and neural network based methods \cite{xie-CVPR-2017} require learning a model for each sequence example.
Different from those methods, we collect a large number of dynamic texture video data and aims at modeling the general distribution of their motions.
Such a model can immediately serve as an editing tool when animating static texture examples.

\section{Proposed Algorithm}

We formulate the video prediction as two phases: flow prediction and flow-to-frame generation. The flow prediction phase, triggered by a noise, directly predicts a set of consecutive flow maps conditioned on the observed first frame. Then the flow-to-frame phase iteratively synthesizes future frames with the previous frame and the corresponding predicted flow map, starting from the first given frame and first predicted flow map.

\begin{figure}[t]
\centering
\begin{tabular}{c@{\hspace{0.005\linewidth}}c}

    \includegraphics[width = .96\linewidth]{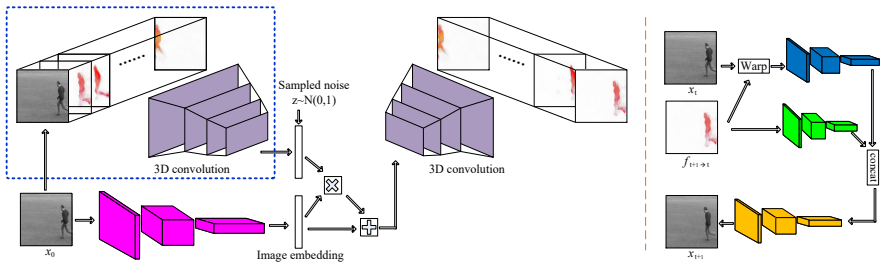} & \\

\end{tabular}
\caption{Architecture of the proposed multi-step prediction network. It consists of a 3D-cVAE (left) for predicting consecutive flows and a \emph{Flow2rgb} model to generate future frame pixels (right). During the testing, the encoder (blue rectangle) of 3D-cVAE is no longer used and we directly sample points from the distribution for predictions.}
\label{fig:framework}
\end{figure}

\subsection{Flow prediction} \label{sec:flow_prediction}

Figure~\ref{fig:framework}(left) illustrates the architecture of our proposed model for predicting consecutive optical flows. Formally, our model is a conditional variational autoencoder~\cite{vae-ICLR-2014,cvae-sohn2015learning} with a spatial-temporal convolutional architecture (3D-cVAE). Given a sequence $X=\{x_i\}^{M}_{0}$ with $x_0$ as the starting frame, we denote the set of consecutive optical flows between adjacent frames in $X$ as $F=\{f_i\}^{M-1}_{0}$. The network is trained to map the observation $F$ (conditioned on $x_0$) to the latent variable $z$ which are likely to reproduce the $F$. In order to avoid training a deterministic model, we produces a distribution over $z$ values, which we sample from before the decoding. Such a variational distribution $q_{\phi}(z\vert x_0,F)$, known as the recognition model in~\cite{cvae-sohn2015learning}, is assumed to be trained to follow a Gaussian distribution $p_{z}(z)$. Given a sampled $z$, the decoder decodes the flow $F$
from the conditional distribution $p_{\theta}(F\vert x_0,z)$. Therefore the whole objective of network training is to maximize the variational lower-bound~\cite{vae-ICLR-2014} of the following negative log-likelihood
function:
\begin{equation} \label{eq:vae}
  \mathcal{L}(x_0, F;\theta,\phi) \approx -\mathcal{D}_{KL} (q_{\phi}(z\vert x_0,F) \vert \vert p_{z}(z)) + \frac{1}{L} \sum^{L}_{1} \log p_{\theta} (F \vert x_0, z),
\end{equation}
where $\mathcal{D}_{KL}$ is the Kullback-Leibler (K-L) divergence and $L$ is the number of samples. 
Maximizing the term at rightmost in (\ref{eq:vae}) is equivalent to minimizing the
L1 distance between the predicted flow and the observed flow. Hence the loss $\mathcal{L}$ consists of a flow reconstruction loss and a K-L divergence loss.
%



\begin{figure}[t]
\centering
\begin{tabular}{c@{\hspace{0.005\linewidth}}c@{\hspace{0.005\linewidth}}c@{\hspace{0.005\linewidth}}c@{\hspace{0.005\linewidth}}c@{\hspace{0.005\linewidth}}c@{\hspace{0.005\linewidth}}c@{\hspace{0.005\linewidth}}c@{\hspace{0.005\linewidth}}c@{\hspace{0.005\linewidth}}c}

    \includegraphics[width = .103\linewidth]{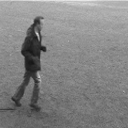} &
    
    \hspace{1pt}\vrule\hspace{1pt}
    
    \includegraphics[width = .103\linewidth]{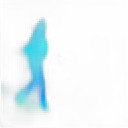} &
    \includegraphics[width = .103\linewidth]{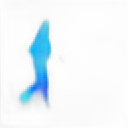} &
    \includegraphics[width = .103\linewidth]{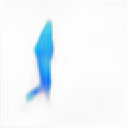} &
    \includegraphics[width = .103\linewidth]{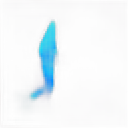} &
    \includegraphics[width = .103\linewidth]{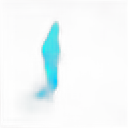} &
    \includegraphics[width = .103\linewidth]{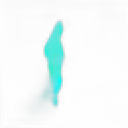} &
    \includegraphics[width = .103\linewidth]{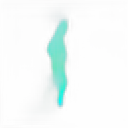} &
    \includegraphics[width = .103\linewidth]{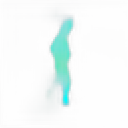} & \\
    
    \includegraphics[width = .103\linewidth]{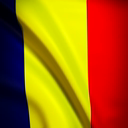} &
    
    \hspace{1pt}\vrule\hspace{1pt}    
    
    \includegraphics[width = .103\linewidth]{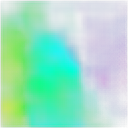} &   
    \includegraphics[width = .103\linewidth]{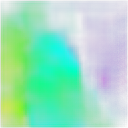} & 
    \includegraphics[width = .103\linewidth]{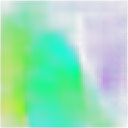} & 
    \includegraphics[width = .103\linewidth]{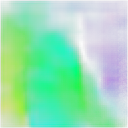} & 
    \includegraphics[width = .103\linewidth]{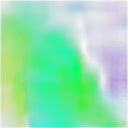} & 
    \includegraphics[width = .103\linewidth]{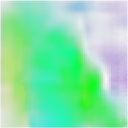} & 
    \includegraphics[width = .103\linewidth]{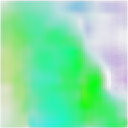} & 
    \includegraphics[width = .103\linewidth]{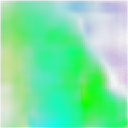} & \\
    
    { t=0  }& { t=1  }& { t=2  }& { t=3  }& { t=4  }& { t=5  }& { t=6  }& { t=7  }& { t=8  } \\
\end{tabular}
\caption{Examples of our multi-step flow prediction. During the testing, by simply sampling a noise from $N\sim (0,1)$, we obtain a set of consecutive flows that describe the future motion field in multiple time steps. Note that since we have a warp operation in the later flow-to-frame step (Section~\ref{sec:frame_generation}) and the backward warping will not result in \emph{holes} in results, we predict the backward flow in this step, i.e., the motion from $x_{t+1}$ to $x_{t}$. 
This is just for convenience and we empirically do not find obvious difference between predicting forward and backward flows.
}
\label{fig:flow_prediction}
\end{figure}

Different from traditional cVAE models~\cite{cvae-sohn2015learning,crossconv-NIPS-2016,walker2016uncertain}, our 3D-cVAE model employs the 3D convolution (purple blocks in Figure~\ref{fig:framework}) which is demonstrated to be well-suited for spatial-temporal feature learning~\cite{c3d-ICCV-2015,videoGAN-NIPS-2016}.
In terms of network architecture, the 3D convolutional network outputs multiple (a volume of) flow maps instead of one, which can be used to predict multiple future frames.
More importantly, the spatial-temporal relationship between adjacent flows are implicitly modeled during the training due to the 3D convolution operations, ensuring that the predicted motions are continuous and reasonable over time.
In order to let the variational distribution $q_{\phi}(z\vert x_0,F)$ conditioned on the starting frame, we stack $x_0$ with each flow map $f_i$ in $F$ as the encoder input. Meanwhile, learning the conditional distribution $p_{\theta}(F\vert x_0,z)$ for flow reconstruction also needs to be conditioned on $x_0$ in the latent space. Therefore, we propose an image encoder (pink blocks in Figure~\ref{fig:framework}) to first map $x_0$ to a latent vector that has the same dimension as $z$. Inspired by the image analogy work~\cite{reed2015deep}, we use a conditioning strategy of combining the multiplication and addition operation, as shown in Figure~\ref{fig:framework}(left).
After we obtain the flow sequence for the future, we proceed to generate the pixel-level full frames.

\subsection{Frame generation} \label{sec:frame_generation}

Given the flow information, a common way to obtain the next frame is warping or pixel copying~\cite{zhou2016view}. However, due to the existence of occlusions, the result is often left with unnecessary pixels inherited from the previous frame. The frame interpolation work~\cite{liu-ICCV-2017} predicts a mask indicating where to copy pixels from previous and next frame. But they require at least two frames to infer the occluded parts. 
Since we only observe one image, it is straightforward to formulate this step as a generation process, meaning that this model can ``imagine'' the appearance of next frame according to the flow and starting frame. The similar idea is also applied in the task of novel view synthesis~\cite{park2017transformation}.

\begin{figure}[t]
\centering
\begin{tabular}{c@{\hspace{0.005\linewidth}}c@{\hspace{0.005\linewidth}}c@{\hspace{0.005\linewidth}}c@{\hspace{0.005\linewidth}}c@{\hspace{0.005\linewidth}}c@{\hspace{0.005\linewidth}}c@{\hspace{0.005\linewidth}}c@{\hspace{0.005\linewidth}}c@{\hspace{0.005\linewidth}}c}

    \includegraphics[width = .103\linewidth]{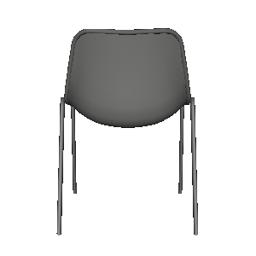} &
    
    \hspace{1pt}\vrule\hspace{1pt}
        
    \includegraphics[width = .103\linewidth]{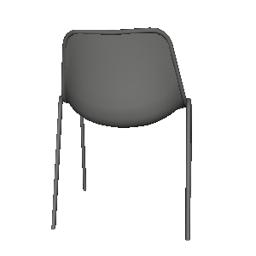} &
    \includegraphics[width = .103\linewidth]{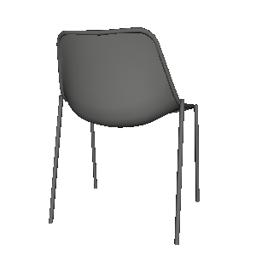} &
    \includegraphics[width = .103\linewidth]{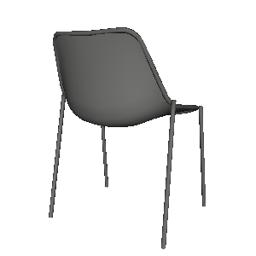} &
    \includegraphics[width = .103\linewidth]{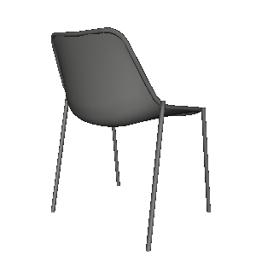} &
    \includegraphics[width = .103\linewidth]{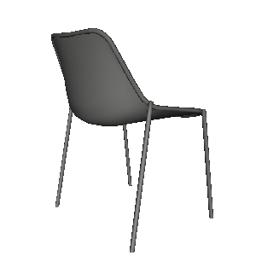} &
    \includegraphics[width = .103\linewidth]{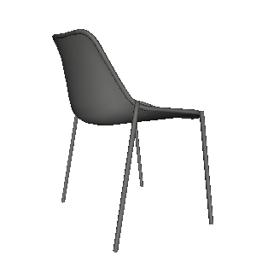} &
    \includegraphics[width = .103\linewidth]{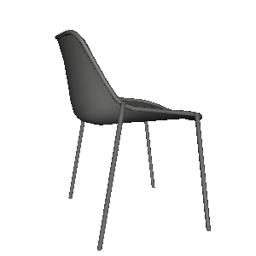} &
    \includegraphics[width = .103\linewidth]{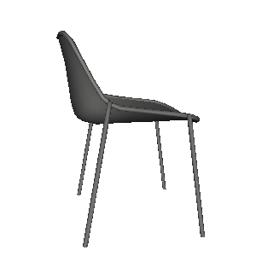} & \\

    \includegraphics[width = .103\linewidth]{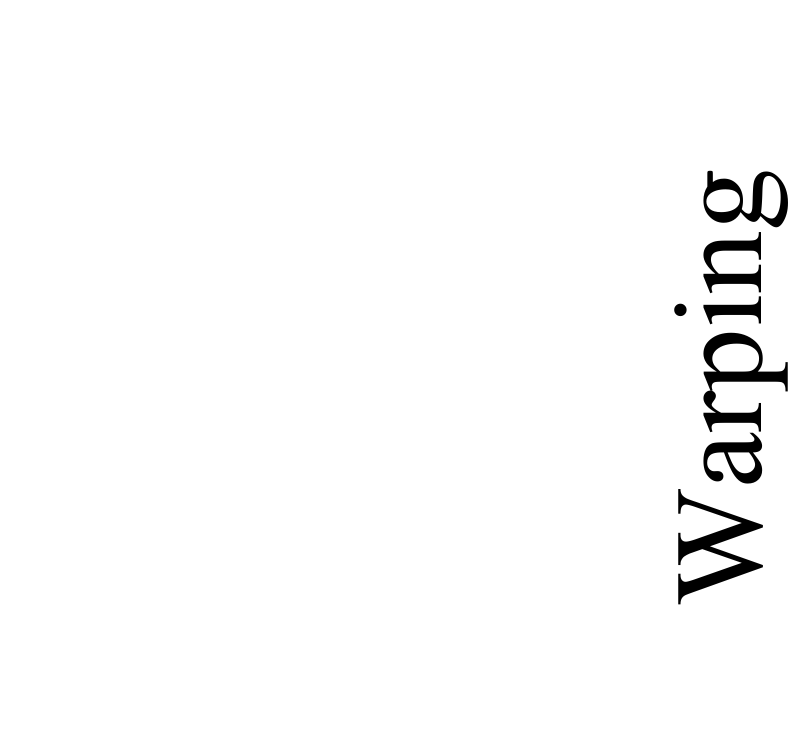} &
    
    \hspace{1pt}\vrule\hspace{1pt}
    
    \includegraphics[width = .103\linewidth]{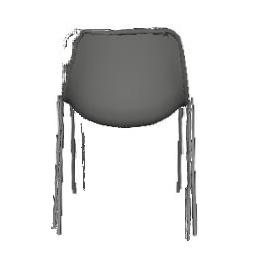} & 
    \includegraphics[width = .103\linewidth]{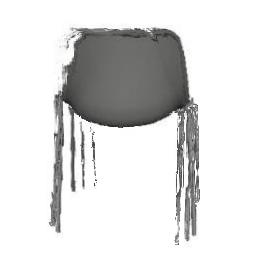} & 
    \includegraphics[width = .103\linewidth]{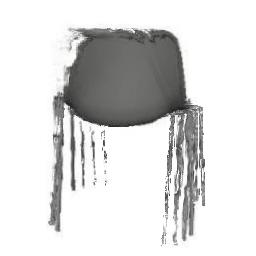} & 
    \includegraphics[width = .103\linewidth]{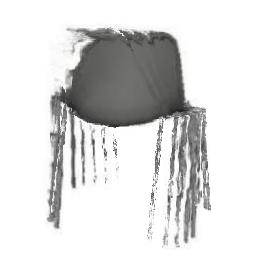} & 
    \includegraphics[width = .103\linewidth]{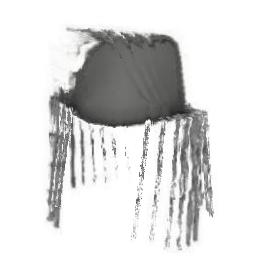} & 
    \includegraphics[width = .103\linewidth]{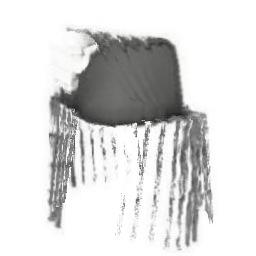} & 
    \includegraphics[width = .103\linewidth]{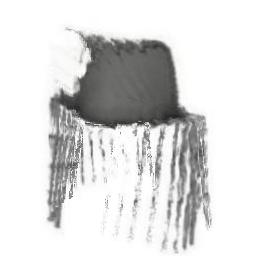} & 
    \includegraphics[width = .103\linewidth]{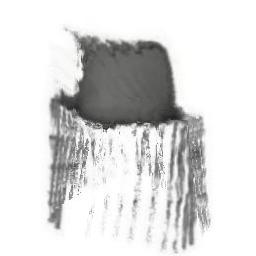} & \\
    
    \includegraphics[width = .103\linewidth]{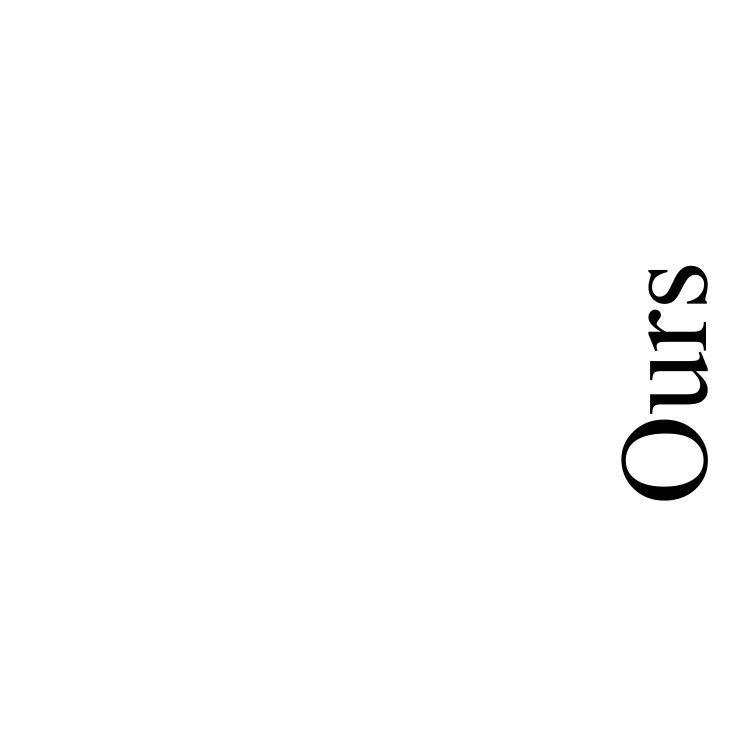} &
    
    \hspace{1pt}\vrule\hspace{1pt}
    
    \includegraphics[width = .103\linewidth]{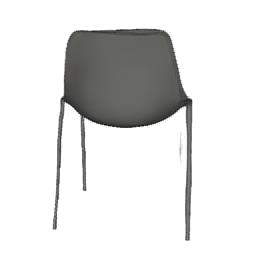} & 
    \includegraphics[width = .103\linewidth]{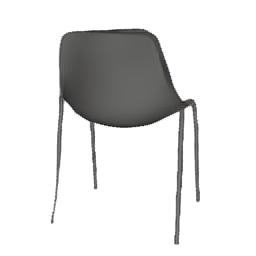} &
    \includegraphics[width = .103\linewidth]{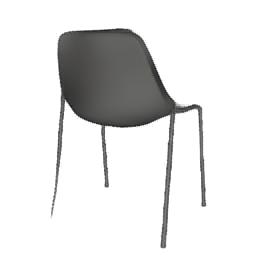} &
    \includegraphics[width = .103\linewidth]{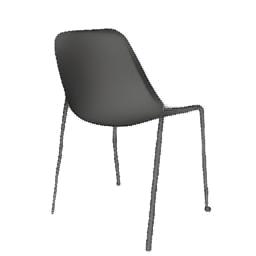} &
    \includegraphics[width = .103\linewidth]{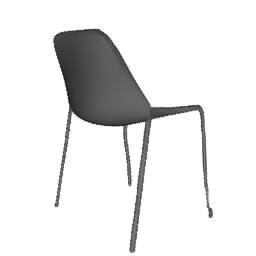} &
    \includegraphics[width = .103\linewidth]{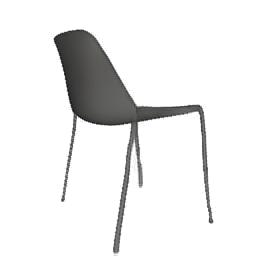} &
    \includegraphics[width = .103\linewidth]{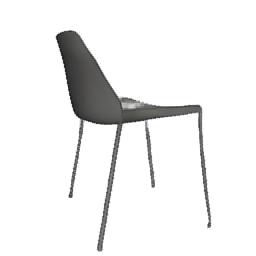} &
    \includegraphics[width = .103\linewidth]{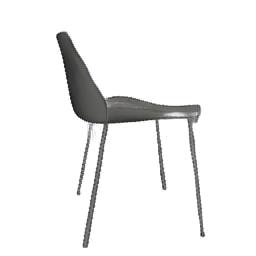} & \\

    { t=0  }& { t=1  }& { t=2  }& { t=3  }& { t=4  }& { t=5  }& { t=6  }& { t=7  }& { t=8  } \\
\end{tabular}
\caption{Comparisons between our \emph{Flow2rgb} model and warping operation, given the first frame and all precomputed flows (between adjacent ground truth frames). Starting from the first frame and first flow, we iteratively run warping or the proposed \emph{Flow2rgb} model based on the previous result and next flow to obtain the sequence. Top: ground truth, Middle: warping results, Bottom: our results.}
\label{fig:chair}
\end{figure}

The architecture of the proposed frame generation model \emph{Flow2rgb} is shown in Figure~\ref{fig:framework}(right). 
Given the input $x_t$ and its optical flow $f_t$ that represents the motion of next time step, the network is trained to generate the next frame $x_{t+1}$. 
%
%
%
Since two adjacent frames often share similar information (especially in the static background regions), in order to let the network focus on learning the difference of two frames, we first warp the $x_t$ based on the flow to get a coarse estimation $\tilde{x}_{t+1}$.
Then we design a Siamese-like~\cite{chopra2005learning} network with the warped frame and the flow as two streams of input. The frame and flow encoders (blue and green blocks) borrow the same architecture of the VGG-19 up to the Relu\_4\_1 layer, and the decoder (yellow blocks) is designed as being symmetrical to the encoder with the nearest neighbor upsampling layer used for enlarging feature maps.
We train the model using a pixel reconstruction loss and a feature loss~\cite{johnson2016perceptual,Doso-NIPS2016-Generation} as shown below:
\begin{equation}\label{reconstruction}
\mathcal{L} = \|\hat{x}_{t+1}-x_{t+1}\|_{2} + \sum^{5}_{K=1} \lambda\|\Phi_{K} (\hat{x}_{t+1}) - \Phi_{K} (x_{t+1})\|_{2}~,
\end{equation}
where $\hat{x}_{t+1}$, $x_{t+1}$ are the network output and ground truth (GT), and $\Phi_K$ is the VGG-19~\cite{VGG-ICLR-2015} encoder that extracts the Relu\_K\_1 features. $\lambda$ is the weight to balance the two losses. This model is learned in an unsupervised manner without human labels.
Note that this is a one-step flow-to-frame model. Since we predict multi-step flows in the flow prediction stage, starting with the first given frame, we iteratively run this model to generate the following frame based on the next flow and previous generated frame.

\begin{figure}[t]
\centering

\begin{tabular}{c@{\hspace{0.005\linewidth}}c@{\hspace{0.005\linewidth}}c@{\hspace{0.005\linewidth}}c}

\includegraphics[width = .48\linewidth]{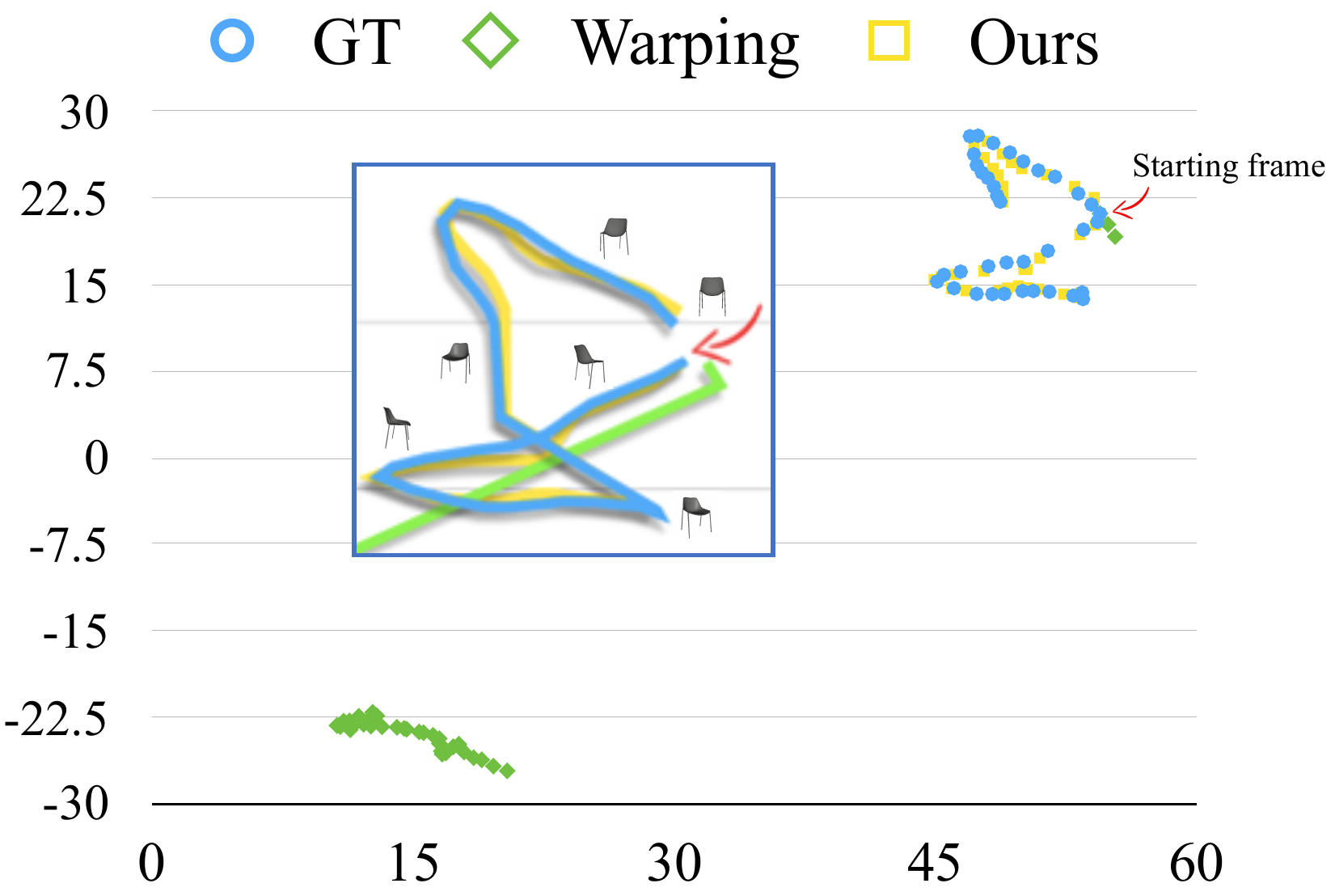} &
\includegraphics[width = .48\linewidth]{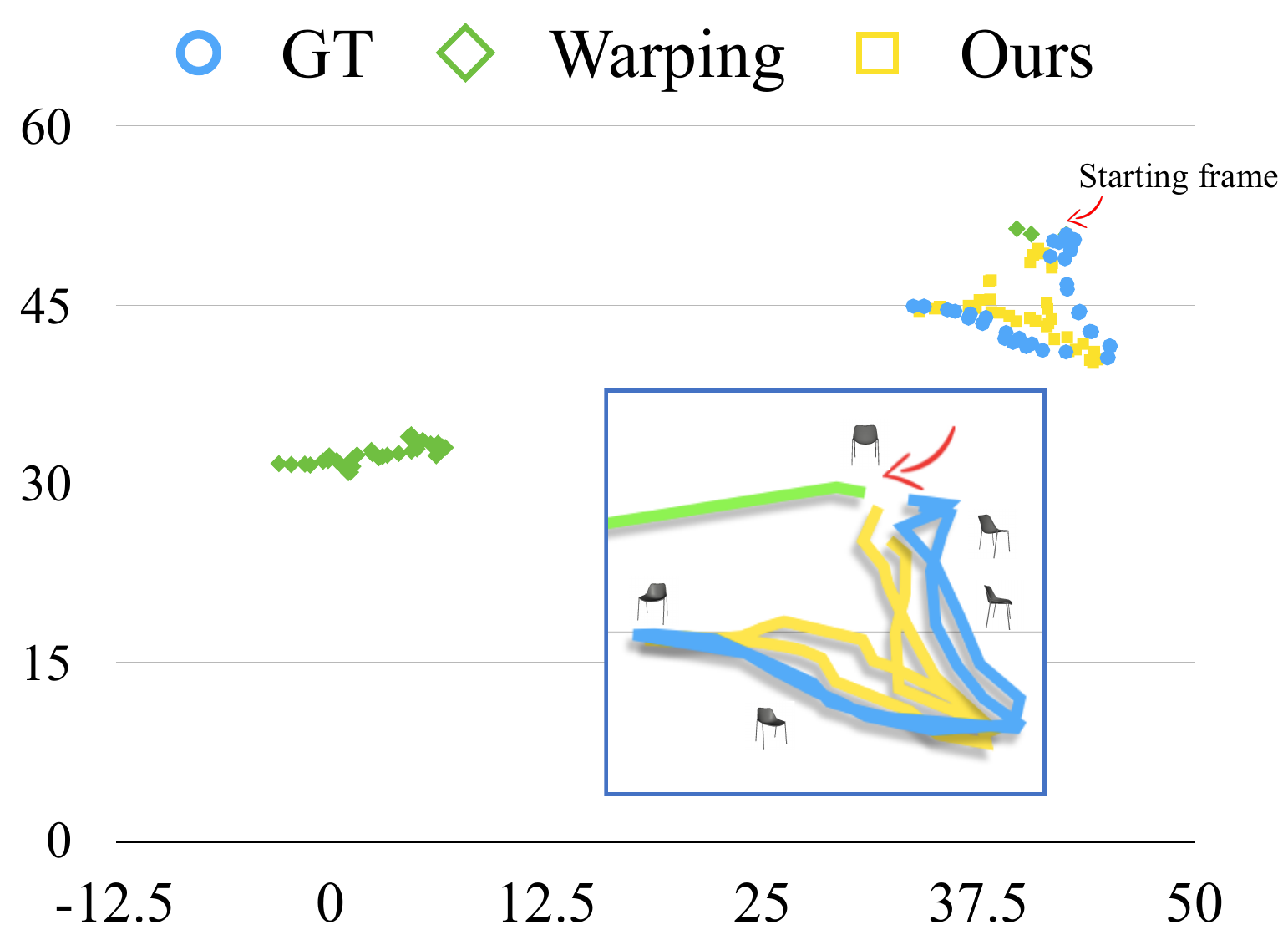} & \\

{(a) VGG-19 pool5} & {(b) VGG-19 fc6} \\

\end{tabular}
\caption{Visualization of sequence (a chair turning around) manifold in deep feature space. Staring from the same frame, each predicted frame of three sequences is visualized as a 2-D point by applying t-SNE~\cite{tsne} on its deep features. The moving average is shown as lines to imply the shape (or trending) of the manifold. For example in (a), the GT rotating chair (blue) follows a ``8'' like manifold in pool5 feature space, which our predicted sequence (yellow) follows closely but the warping sequence (green) deviates much further.
}
\label{fig:embedding_manifold}
\end{figure}

We show the effectiveness of our \emph{Flow2rgb} model in Figure~\ref{fig:chair} with an example of chair rotating sequence~\cite{chair-CVPR-2015}. To verify the frame generation phase alone, we assume that the flows are already available (computed by~\cite{spynet-CVPR-2017}). Then given the first frame and future flows, the second row of Figure~\ref{fig:chair} shows the iterative warping results where the chair legs are repeatedly copied in future frames as the warping is unable to depict the right appearance of chair in difference views.
In contrast, our model iteratively generates the occluded parts and removed unnecessary parts in the previous frame according to the flow at each time step.
As claimed in~\cite{chair-CVPR-2015}, the deep embeddings of objects under consecutively changing views often follow certain manifold in feature space. If we interpret this changing view as a type of rotating motion, our predicted results for different views also needs to stay close to the manifold shape of the GT sequence. 
We demonstrate this by extracting the VGG-19~\cite{VGG-ICLR-2015} features of each predicted frame, mapping it to a 2-D point through t-SNE~\cite{tsne}, and visualizing it in Figure~\ref{fig:embedding_manifold}. 
It clearly shows that our predictions follows closely with the manifold of the GT sequence, while the warping drives the predictions to deviate from the GT further and further.

\section{Experimental Results}

In this section, we first discuss the experimental settings and implementation details. 
We then present qualitative and quantitative comparisons between the proposed algorithm and several competing algorithms. 
Finally, we analyze the diversity issue in uncertainty modeling.

\paragraph{\bf Datasets.} We mainly evaluate our algorithm on three datasets. The first one is the KTH dataset~\cite{KTH-ICPR-2004} which is a human action video dataset that consists of six types of action and totally 600 videos. It represents the movement of articulated objects. Same as in~\cite{mcnet-ICLR-2017,drnet-NIPS-2017}, we use person 1-16 for training and 17-25 for testing. We also collect another two datasets from online websites,
i.e., the \emph{WavingFlag} and \emph{FloatingCloud}. These two datasets represents dynamic texture videos where motions may bring the shape changes on dynamic patterns. The \emph{WavingFlag} dataset contains 341 videos of 80K+ frames and the \emph{FloatingCloud} dataset has 415 videos of 150K+ frames in total. In each dataset, we randomly split all videos into the training (4/5) and testing (1/5) set.
%

\paragraph{\bf Implementation details.}~Given the starting frame $x_0$, our algorithm predicts the future in next $M=16$ time steps. Each frame is resized to 128$\times$128 in experiments. Similar to~\cite{walker-ICCV-2015,gao-2017im2flow}, we employ an existing optical flow estimator SPyNet~\cite{spynet-CVPR-2017} to obtain flows between GT frames for training the 3D-cVAE. As described in Section~\ref{sec:flow_prediction}, we stack $x_0$ with each flow map $f_i$ in $F$. Thus during the training, the input cube to the 3D-cVAE is of size $16\times 5 \times 128\times 128$ where $5=2+3$ (2-channel flow and 3-channel RGB). The dimension of the latent variable $z$ in the bottle neck is set as 2000. 
%
%
Another important factor for a successful network training is to normalize the flow roughly to (0,1) before feeding it into the network, ensuring pixel values of both flows and RGB frames are within the similar range.
Since the \emph{Flow2rgb} model can be an independent module for motion transfer with known flows, we train the 3D-cVAE and \emph{Flow2rgb} model separately in experiments.

\afterpage{\clearpage}
\begin{figure}[t]
\centering
\begin{tabular}{c@{\hspace{0.005\linewidth}}c@{\hspace{0.005\linewidth}}c@{\hspace{0.005\linewidth}}c@{\hspace{0.005\linewidth}}c@{\hspace{0.005\linewidth}}c@{\hspace{0.005\linewidth}}c@{\hspace{0.005\linewidth}}c@{\hspace{0.005\linewidth}}c@{\hspace{0.005\linewidth}}c}

    \includegraphics[width = .103\linewidth]{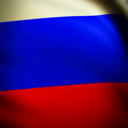} &
    
    \hspace{1pt}\vrule\hspace{1pt}
    
    \includegraphics[width = .103\linewidth]{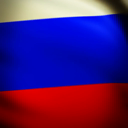} &
    \includegraphics[width = .103\linewidth]{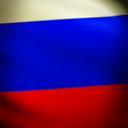} &
    \includegraphics[width = .103\linewidth]{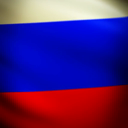} &
    \includegraphics[width = .103\linewidth]{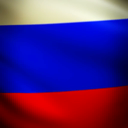} &
    \includegraphics[width = .103\linewidth]{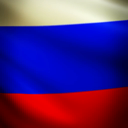} &
    \includegraphics[width = .103\linewidth]{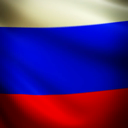} &
    \includegraphics[width = .103\linewidth]{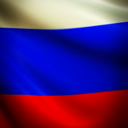} &
    \includegraphics[width = .103\linewidth]{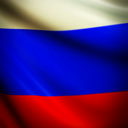} & \\

    \includegraphics[height = .103\linewidth, width = .103\linewidth]{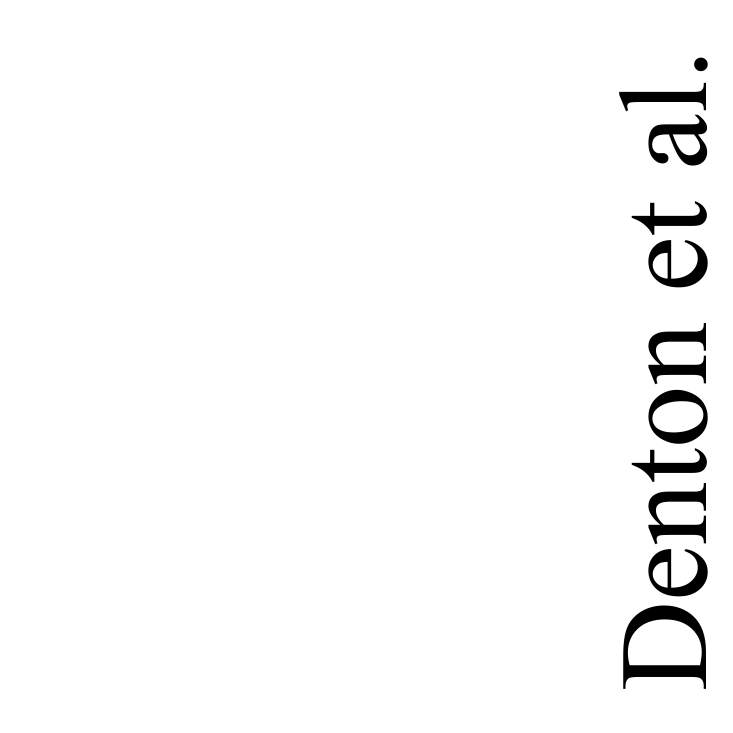} &
    
    \hspace{1pt}\vrule\hspace{1pt}
    
    \includegraphics[width = .103\linewidth]{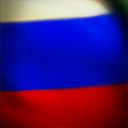} &
    \includegraphics[width = .103\linewidth]{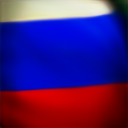} &
    \includegraphics[width = .103\linewidth]{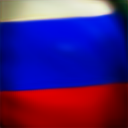} &
    \includegraphics[width = .103\linewidth]{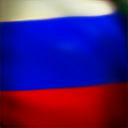} &
    \includegraphics[width = .103\linewidth]{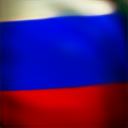} &
    \includegraphics[width = .103\linewidth]{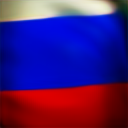} &
    \includegraphics[width = .103\linewidth]{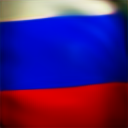} &
    \includegraphics[width = .103\linewidth]{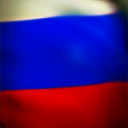} & \\

    \includegraphics[width = .103\linewidth]{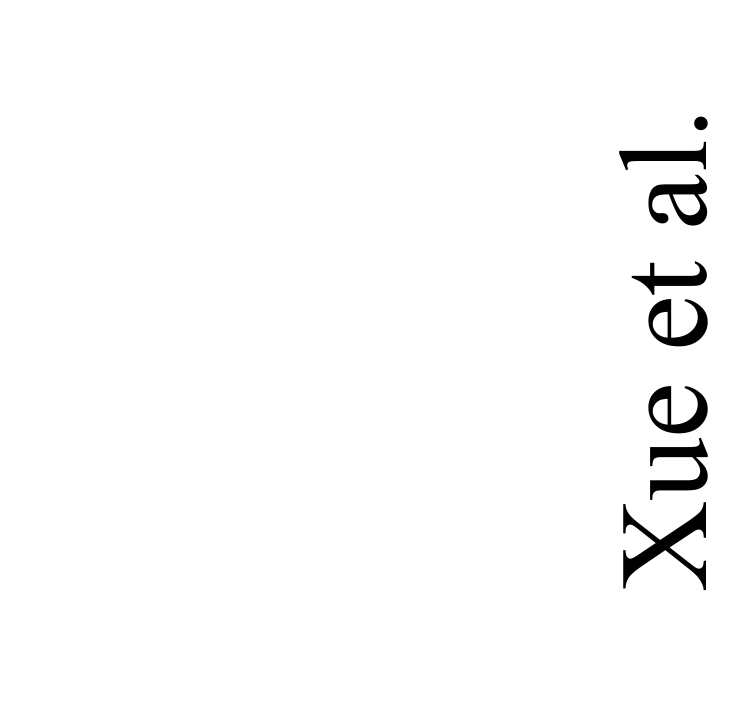} &
    
    \hspace{1pt}\vrule\hspace{1pt}
    
    \includegraphics[width = .103\linewidth]{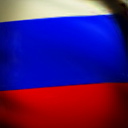} &
    \includegraphics[width = .103\linewidth]{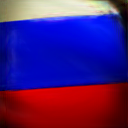} &
    \includegraphics[width = .103\linewidth]{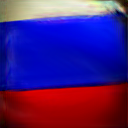} &
    \includegraphics[width = .103\linewidth]{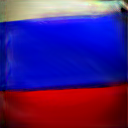} &
    \includegraphics[width = .103\linewidth]{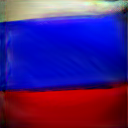} &
    \includegraphics[width = .103\linewidth]{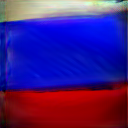} &
    \includegraphics[width = .103\linewidth]{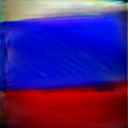} &
    \includegraphics[width = .103\linewidth]{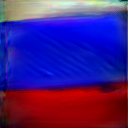} & \\

    \includegraphics[width = .103\linewidth]{figs/visual_result/ours.pdf} &
    
    \hspace{1pt}\vrule\hspace{1pt}
    
    \includegraphics[width = .103\linewidth]{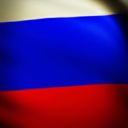} &
    \includegraphics[width = .103\linewidth]{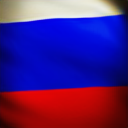} &
    \includegraphics[width = .103\linewidth]{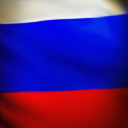} &
    \includegraphics[width = .103\linewidth]{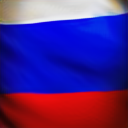} &
    \includegraphics[width = .103\linewidth]{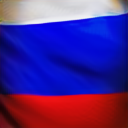} &
    \includegraphics[width = .103\linewidth]{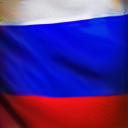} &
    \includegraphics[width = .103\linewidth]{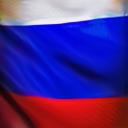} &
    \includegraphics[width = .103\linewidth]{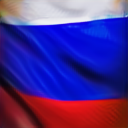} & \\

    \includegraphics[width = .103\linewidth]{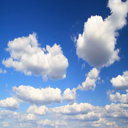} &
    
    \hspace{1pt}\vrule\hspace{1pt}
    
    \includegraphics[width = .103\linewidth]{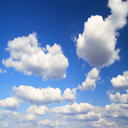} &
    \includegraphics[width = .103\linewidth]{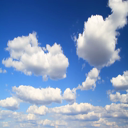} &
    \includegraphics[width = .103\linewidth]{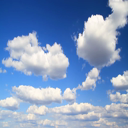} &
    \includegraphics[width = .103\linewidth]{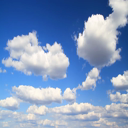} &
    \includegraphics[width = .103\linewidth]{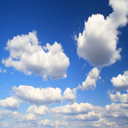} &
    \includegraphics[width = .103\linewidth]{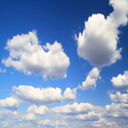} &
    \includegraphics[width = .103\linewidth]{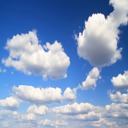} &
    \includegraphics[width = .103\linewidth]{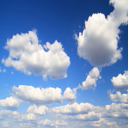} & \\

    \includegraphics[width = .103\linewidth]{figs/visual_result/denton.pdf} &
    
    \hspace{1pt}\vrule\hspace{1pt}
    
    \includegraphics[width = .103\linewidth]{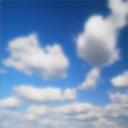} &
    \includegraphics[width = .103\linewidth]{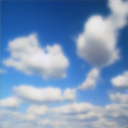} &
    \includegraphics[width = .103\linewidth]{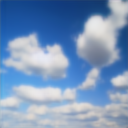} &
    \includegraphics[width = .103\linewidth]{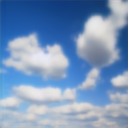} &
    \includegraphics[width = .103\linewidth]{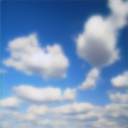} &
    \includegraphics[width = .103\linewidth]{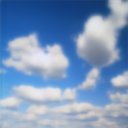} &
    \includegraphics[width = .103\linewidth]{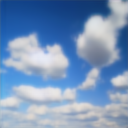} &
    \includegraphics[width = .103\linewidth]{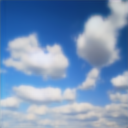} & \\

    \includegraphics[width = .103\linewidth]{figs/visual_result/xue.pdf} &
    
    \hspace{1pt}\vrule\hspace{1pt}
    
    \includegraphics[width = .103\linewidth]{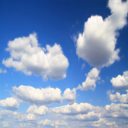} &
    \includegraphics[width = .103\linewidth]{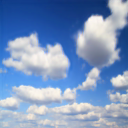} &
    \includegraphics[width = .103\linewidth]{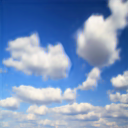} &
    \includegraphics[width = .103\linewidth]{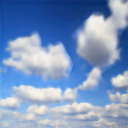} &
    \includegraphics[width = .103\linewidth]{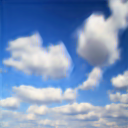} &
    \includegraphics[width = .103\linewidth]{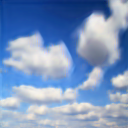} &
    \includegraphics[width = .103\linewidth]{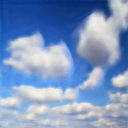} &
    \includegraphics[width = .103\linewidth]{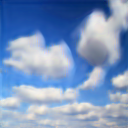} & \\

    \includegraphics[width = .103\linewidth]{figs/visual_result/ours.pdf} &
    
    \hspace{1pt}\vrule\hspace{1pt}
    
    \includegraphics[width = .103\linewidth]{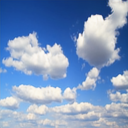} &
    \includegraphics[width = .103\linewidth]{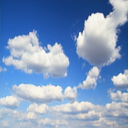} &
    \includegraphics[width = .103\linewidth]{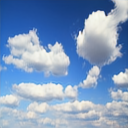} &
    \includegraphics[width = .103\linewidth]{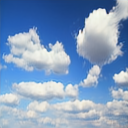} &
    \includegraphics[width = .103\linewidth]{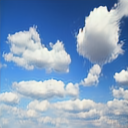} &
    \includegraphics[width = .103\linewidth]{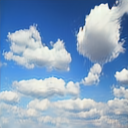} &
    \includegraphics[width = .103\linewidth]{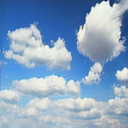} &
    \includegraphics[width = .103\linewidth]{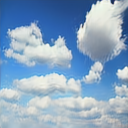} & \\
    
    \includegraphics[width = .103\linewidth]{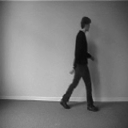} &
    
    \hspace{1pt}\vrule\hspace{1pt}
    
    \includegraphics[width = .103\linewidth]{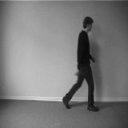} &
    \includegraphics[width = .103\linewidth]{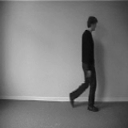} &
    \includegraphics[width = .103\linewidth]{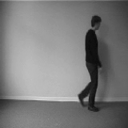} &
    \includegraphics[width = .103\linewidth]{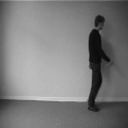} &
    \includegraphics[width = .103\linewidth]{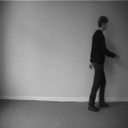} &
    \includegraphics[width = .103\linewidth]{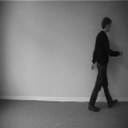} &
    \includegraphics[width = .103\linewidth]{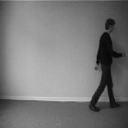} &
    \includegraphics[width = .103\linewidth]{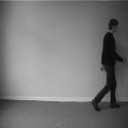} & \\

    \includegraphics[width = .103\linewidth]{figs/visual_result/denton.pdf} &
    
    \hspace{1pt}\vrule\hspace{1pt}
    
    \includegraphics[width = .103\linewidth]{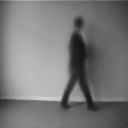} &
    \includegraphics[width = .103\linewidth]{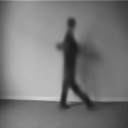} &
    \includegraphics[width = .103\linewidth]{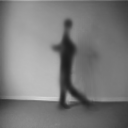} &
    \includegraphics[width = .103\linewidth]{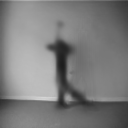} &
    \includegraphics[width = .103\linewidth]{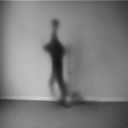} &
    \includegraphics[width = .103\linewidth]{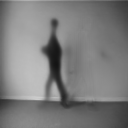} &
    \includegraphics[width = .103\linewidth]{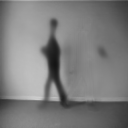} &
    \includegraphics[width = .103\linewidth]{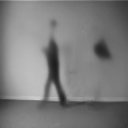} & \\

    \includegraphics[width = .103\linewidth]{figs/visual_result/xue.pdf} &
    
    \hspace{1pt}\vrule\hspace{1pt}
    
    \includegraphics[width = .103\linewidth]{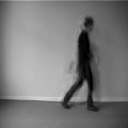} &
    \includegraphics[width = .103\linewidth]{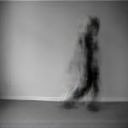} &
    \includegraphics[width = .103\linewidth]{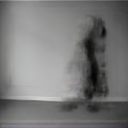} &
    \includegraphics[width = .103\linewidth]{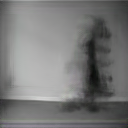} &
    \includegraphics[width = .103\linewidth]{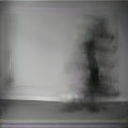} &
    \includegraphics[width = .103\linewidth]{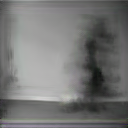} &
    \includegraphics[width = .103\linewidth]{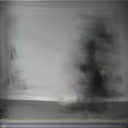} &
    \includegraphics[width = .103\linewidth]{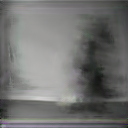} & \\

    \includegraphics[width = .103\linewidth]{figs/visual_result/ours.pdf} &
    
    \hspace{1pt}\vrule\hspace{1pt}
    
    \includegraphics[width = .103\linewidth]{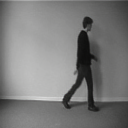} &
    \includegraphics[width = .103\linewidth]{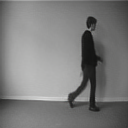} &
    \includegraphics[width = .103\linewidth]{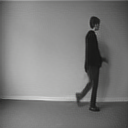} &
    \includegraphics[width = .103\linewidth]{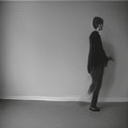} &
    \includegraphics[width = .103\linewidth]{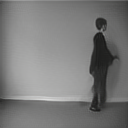} &
    \includegraphics[width = .103\linewidth]{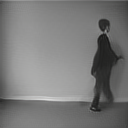} &
    \includegraphics[width = .103\linewidth]{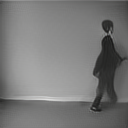} &
    \includegraphics[width = .103\linewidth]{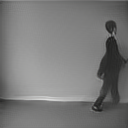} & \\

    { t=0  }& { t=1  }& { t=3  }& { t=5  }& { t=7  }& { t=9  }& { t=11  }& { t=13  }& { t=15  } \\
\end{tabular}
\caption{Visual comparisons of different prediction algorithms. Top left: the starting frame. From top to bottom in example: GT, Denton \etal~\cite{drnet-NIPS-2017}, Xue \etal~\cite{crossconv-NIPS-2016}, Ours. The GT sequence provides a sense of motion rightness, while the predicted sequence is unnecessary to be exactly the same with GT.}
\label{fig:visual_result}
\end{figure}

\paragraph{\bf Evaluations.}~Different prediction algorithms have their unique settings and assumptions. For example, Mathieu \etal~\cite{mathieu-ICLR-2016} requires four frames stacked together as the input. Villegas \etal~\cite{mcnet-ICLR-2017} ask for feeding the image difference (at least two frames). Their following work~\cite{ruben-ICML-2017}, though based on one frame, additionally needs multiple historical human pose maps to start the prediction.
For fair comparisons, we mainly select prediction methods~\cite{drnet-NIPS-2017,crossconv-NIPS-2016} that accept one single image as the only input to compare.
The work of~\cite{drnet-NIPS-2017} represents the typical recursive prediction pipeline, which builds upon a fully-connected long short-term memory (FC-LSTM) layer for predictions. Their model is originally trained and tested by observing multiple frames. Here we change their setting to one-frame observance in order to be consistent with our setting.
The work of~\cite{crossconv-NIPS-2016} is the typical one-step prediction method based on one given frame. To get multi-frame predictions, we train their model and iteratively test it to get the next prediction based on the previous prediction. 
%
%

In Figure~\ref{fig:visual_result}, we provide a visual comparison
between the proposed algorithm and~\cite{drnet-NIPS-2017,crossconv-NIPS-2016}. 
In~\cite{drnet-NIPS-2017}, a pre-trained and disentangled \emph{pose} embedding is employed to keep predicting the pose of the next frame through a FC-LSTM module. 
%
\Yijun{ 
%
%
For articulated objects, the pose is often compact and in low dimensions, which is relatively easier to handle with a single LSTM module.
However, for dynamic textures (e.g., flag, cloud) where all pixels are likely to move, the global pose becomes complex and is no longer a low-dimensional structure representation. Therefore the capacity of recursive models is not enough to capture the spatial and temporal variation trend at the same time.
}
The first two examples in Figure~\ref{fig:visual_result} show that the flag and cloud in predicted frames are nearly static. 
Meanwhile, the pose only describes the static structure of the object in the current frame and cannot tell as much information as the flow about the next-step motion. 
In the third example of Figure~\ref{fig:visual_result}, it is obvious that the human is walking to the right. But the results of~\cite{drnet-NIPS-2017} show that the human is going in a reverse direction.   
Moreover, since they directly predict frame pixels and use the reconstruction loss only, their results are relatively blurry.
In~\cite{crossconv-NIPS-2016}, as they only predict the next one frame, the motion is often clear in the second frame. But after we keep predicting the following frame using the previous predicted frame, the motion gradually disappears and the quality of results degrades fast during a few steps. Moreover, they choose to predict the image difference which only shows global image changes but does not capture how each pixel will move to its corresponding one in the next frame.
In contrast, our results show more continuous and reasonable motion, reflected by better generated full frames. For example, in the first flag example, the starting frame indicates that the fold on top right will disappear and the fold at bottom left will bring bigger folds. Our predicted sequence presents the similar dynamics as what happens in the GT sequence, which makes it look more realistic.
%

\begin{figure}[t]
\centering

\begin{tabular}{c@{\hspace{0.005\linewidth}}c@{\hspace{0.005\linewidth}}c@{\hspace{0.005\linewidth}}c}

\includegraphics[width = .48\linewidth]{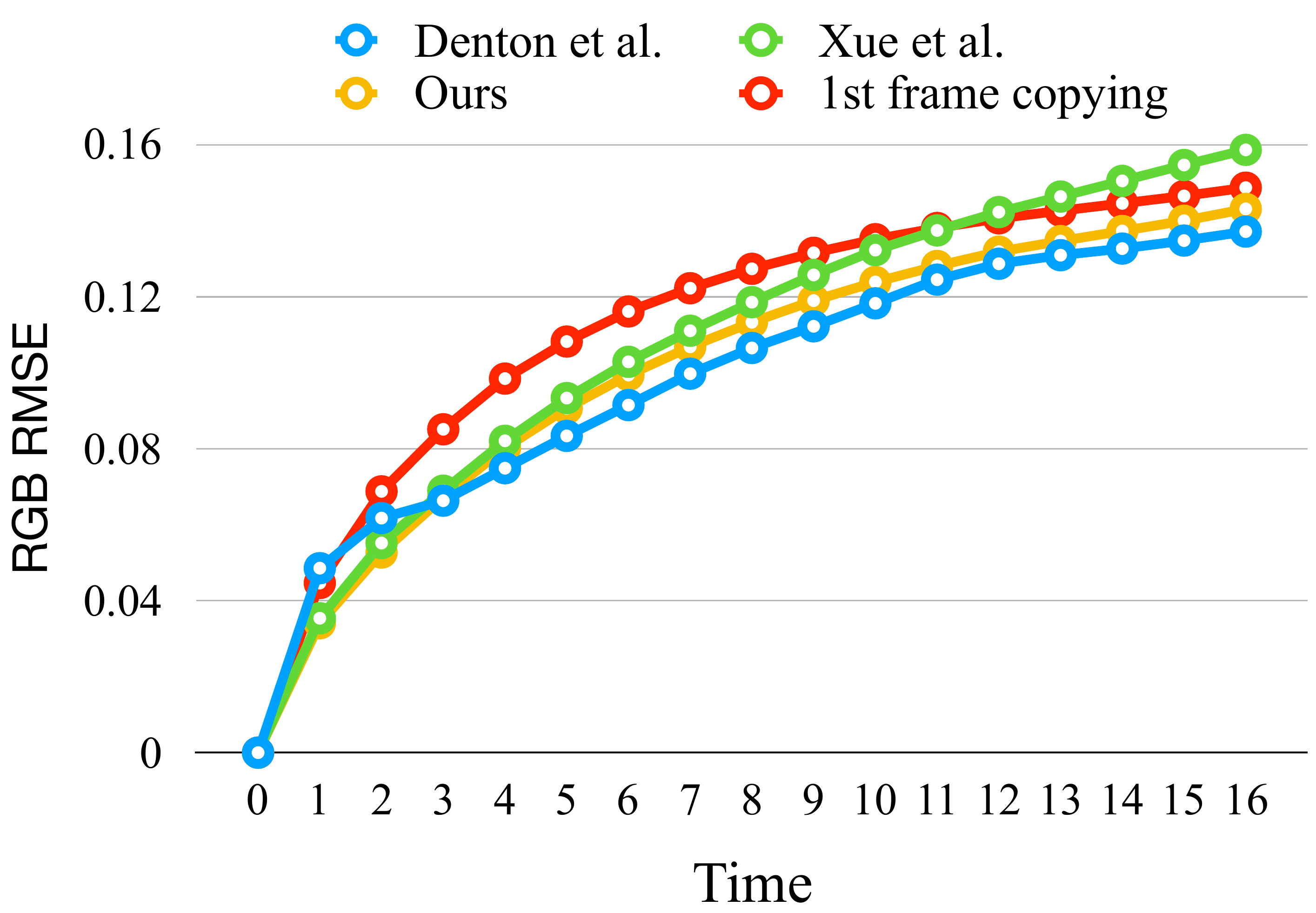} & 
\includegraphics[width = .48\linewidth]{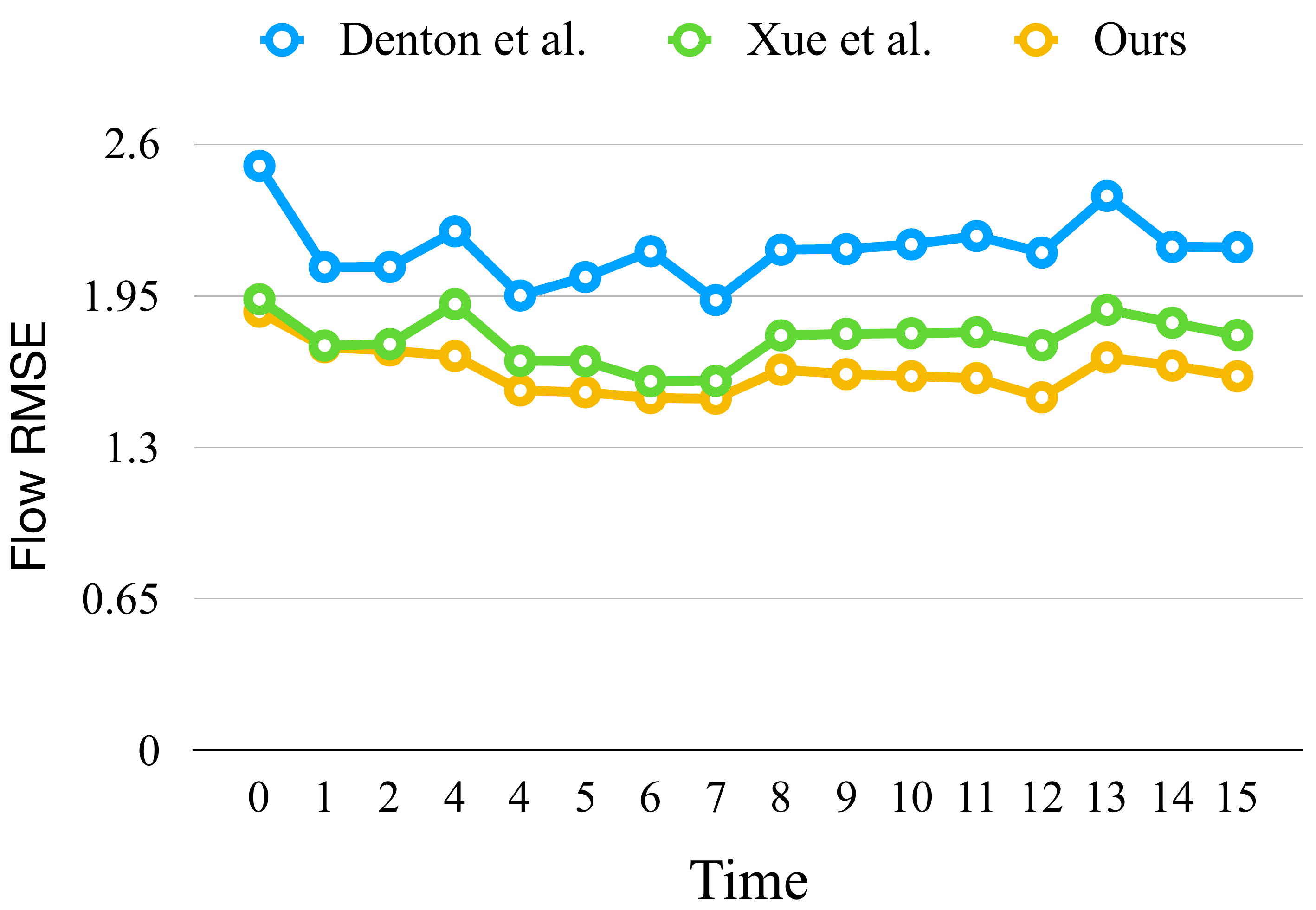} & \\
{(a) RMSE on frames} & {(b) RMSE on flows} \\
\includegraphics[width = .48\linewidth]{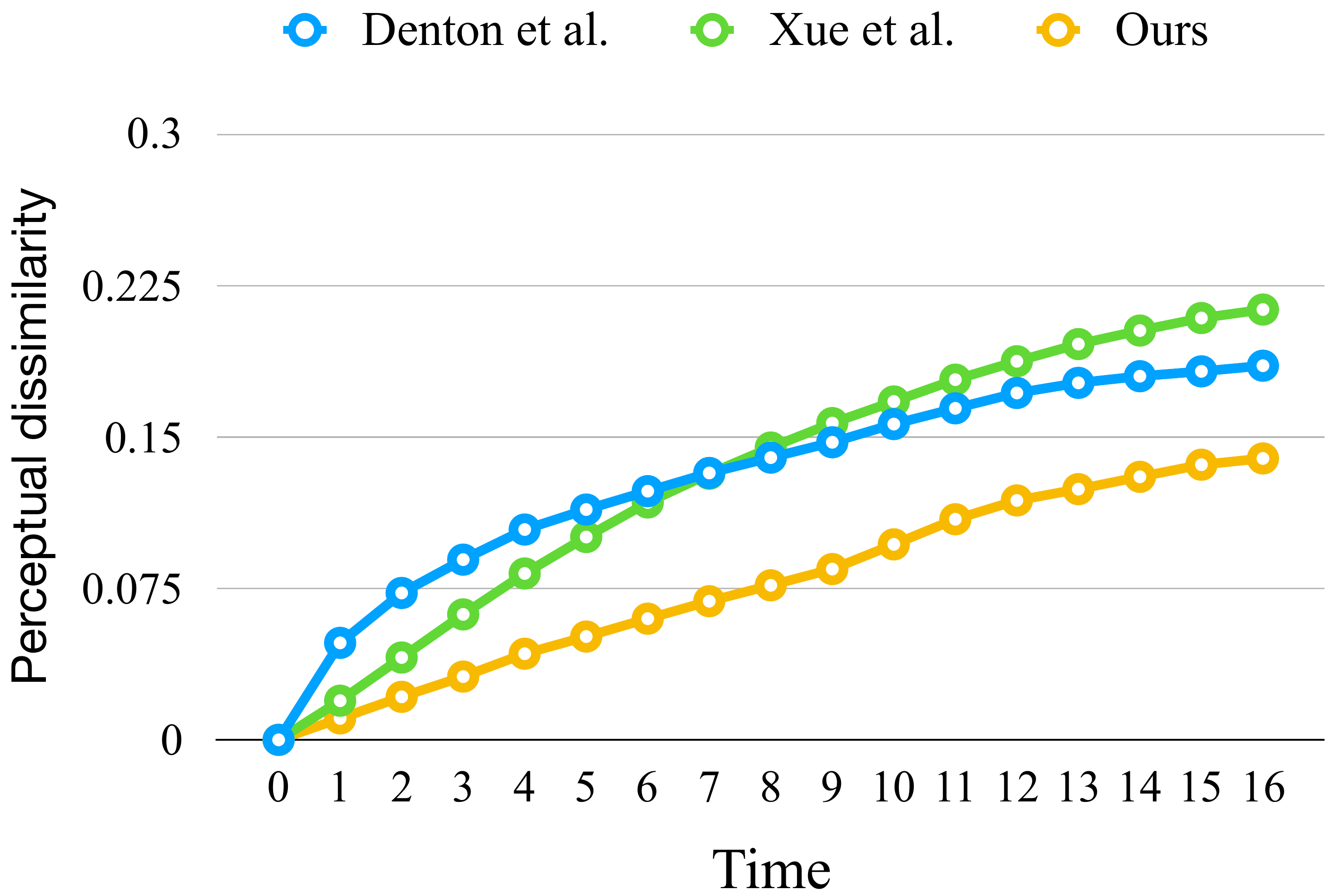} &  
\includegraphics[width = .48\linewidth]{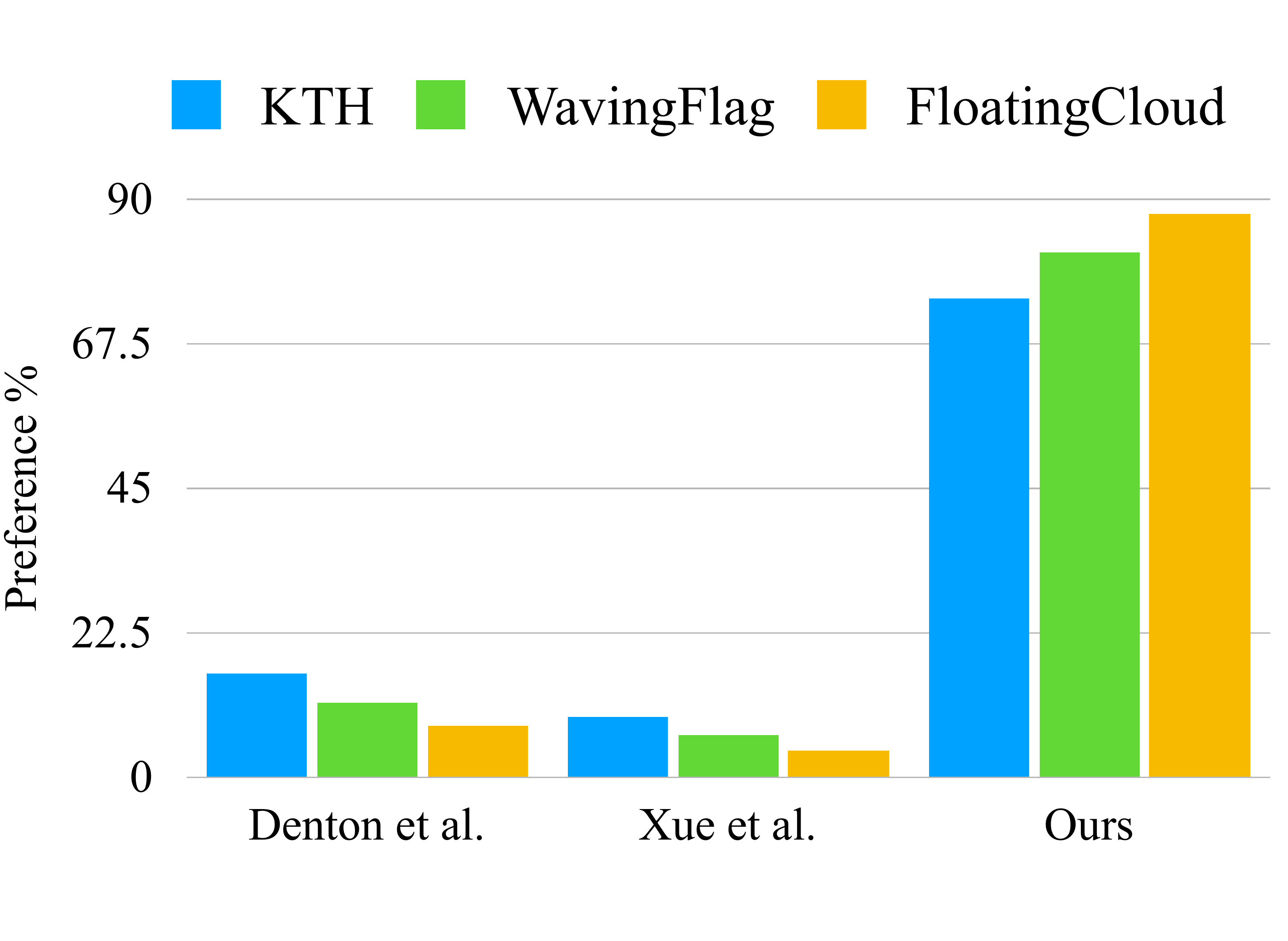} & \\

{(c) Perceptual metric~\cite{zhang2018perceptual} on frames} & {(d) Which sequence looks more realistic? } \\

\end{tabular}
\caption{Quantitative evaluations of different prediction algorithms. We start from the per-pixel metrics (e.g., RMSE) and gradually take human perception into consideration. Our method 
achieves the best performance under metrics (b)-(d).}
\label{fig:quantitative}
\end{figure}

\Yijun{ 
%
%
We also quantitatively evaluate these prediction methods using three different metrics, i.e., the root-mean-square error (RMSE), perceptual similarity~\cite{zhang2018perceptual}, and user preference.
The RMSE is the classic per-pixel metric which measures the spatial correspondence without considering any high-level semantics and is often easily favored by smooth results.  
Based on this observation, the recent work of ~\cite{zhang2018perceptual} proposes a perceptual similarity metric by using deep network embeddings. It is demonstrated to agree with human perceptions better. 
Lastly, we directly ask the feedback from users by conducting user studies to understand their preference towards the predicted results by different algorithms.
}

We start with the traditional RMSE to compute the difference between predicted sequence and GT sequence frame-by-frame and show the result in Figure~\ref{fig:quantitative}(a).
To understand how effective these prediction methods are, we design a simple baseline by copying the given frame as multi-step predictions. However, we do not observe obvious difference among all these methods. While the prediction from one single image is originally ambiguous, the GT sequence can be regarded as just one possibility of the future. The trending of motion may be similar but the resulted images can be significantly different in pixel-level. But the RMSE metric is actually very sensitive to the pixel spatial mismatch. Similar observations are also found in~\cite{drnet-NIPS-2017,zhang2018perceptual}. That is why all these methods, when comparing with the GT sequence, shows the similar RMSE results. Therefore, instead of measuring the RMSE on frames, we turn to measure the RMSE on optical flows because the optical flow represents whether the motion field is predicted similarly or not. 
We compute the flow maps between adjacent frames of the GT sequence and other predicted sequences using the SPyNet~\cite{spynet-CVPR-2017} and show the RMSE results in Figure~\ref{fig:quantitative}(b). Now the difference becomes more clear and our method achieves the lowest RMSE results, meaning that our prediction is the closest to the GT in terms of the predicted motions.

However, the evaluation of prediction results still need to take human perception into consideration in order to determine whether sequences look as realistic as the GT sequence. Therefore we turn to the perceptual similarity metric~\cite{zhang2018perceptual}. 
We use the Alex-Net~\cite{krizhevsky2012imagenet} for feature extraction and measure the similarity between predicted sequence and GT sequence frame-by-frame. Since this metric is obtained by computing feature distances, we denote it as perceptual dissimilarity so that small values means being more similar.
The results in Figure~\ref{fig:quantitative}(c) show that the proposed method outperforms other algorithms with an even larger margin than that in Figure~\ref{fig:quantitative}(b), which means that the predicted sequence of our method is perceptually more similar to the GT sequence.

Finally, we conduct the user study to get the feedback from human subjects on judging different predicted results. We prepare 30 starting frames (10 from each dataset) and generated 30 sequences (16-frame) for each method.
For each subject, we randomly select 15 sets of sequences predicted by three methods. For each starting frame, the three predicted sequences are displayed side-by-side in random order. 
Each subject is asked to vote one sequence that looks most realistic for each starting frame. We finally collect 900 votes from 60 users and report the results (in percentage) in Figure~\ref{fig:quantitative}(d). The study results clearly show that the proposed method receives the most votes for more realistic predictions among all three categories. Both Figure~\ref{fig:quantitative}(c) and (d) indicate that the proposed method performs favorably against~\cite{drnet-NIPS-2017,crossconv-NIPS-2016} in terms of perceptual quality.

\begin{figure}[t]
\centering
\begin{tabular}{c@{\hspace{0.005\linewidth}}c@{\hspace{0.005\linewidth}}c}

    \includegraphics[width = .59\linewidth]{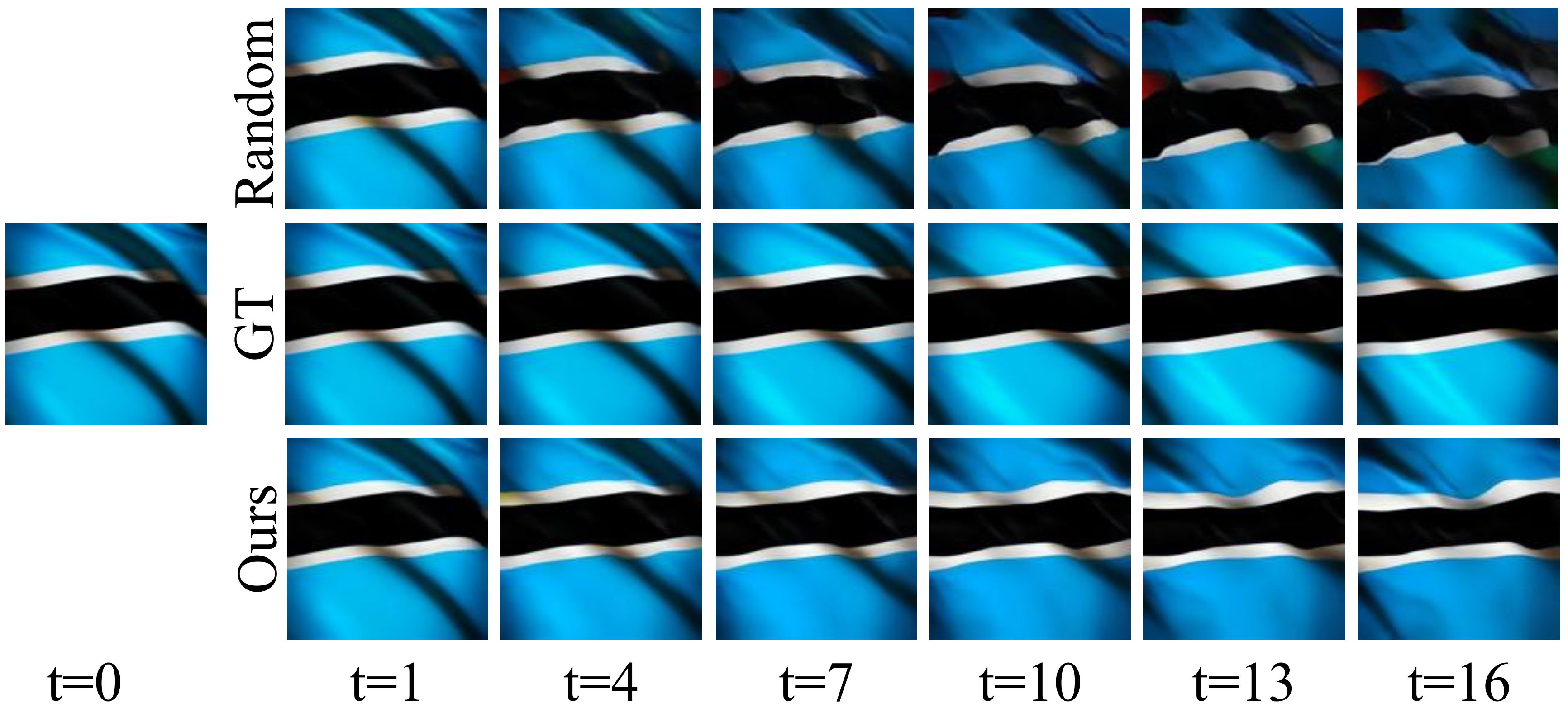} & 
    \includegraphics[width = .37\linewidth]{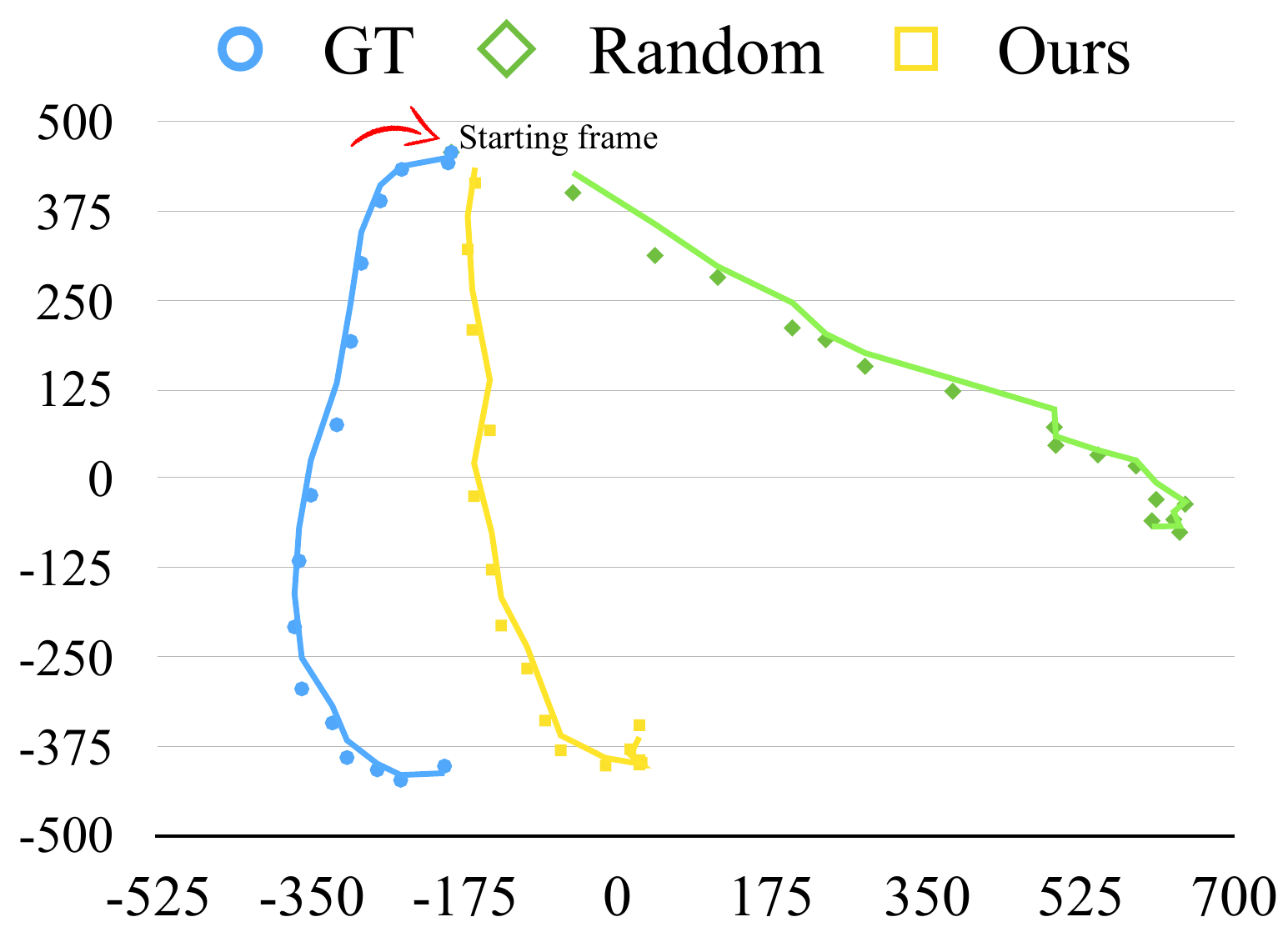} \\
    
{(a) Visual comparisons of an exemplary sequence} & {(b) VGG-19 pool5} \\

\end{tabular}
\caption{Comparison with a naive baseline which transfers a random motion field.
(b) The
GT sequence follows a ``C" like manifold in pool5 feature space, which our
prediction follows closely but the random prediction deviates
much further.
}
\label{fig:random}
\end{figure}

\paragraph{\bf Random motion.}~We also compare with a naive approach which uses random flow maps (e.g., sampling from the Gaussian distribution $N(0,2)$ for each pixel).
We apply the proposed \emph{flow2rgb} model to both random and the learned motions by our method to generate frames.
Figure~\ref{fig:random}(a) shows one example. 
In Figure~\ref{fig:random}(b), we visualize the manifold of predicted sequences in the deep feature
space using the t-SNE scheme (as did in Figure~\ref{fig:embedding_manifold}). 
Both demonstrate that the learned motion generates much better results than those by the random motion, as the naive approach neither models the motion distribution nor considers the temporal relationship between frames.

\begin{figure}[t]
\centering

\begin{tabular}{c@{\hspace{0.005\linewidth}}c@{\hspace{0.005\linewidth}}c@{\hspace{0.005\linewidth}}c}

\includegraphics[width = .48\linewidth]{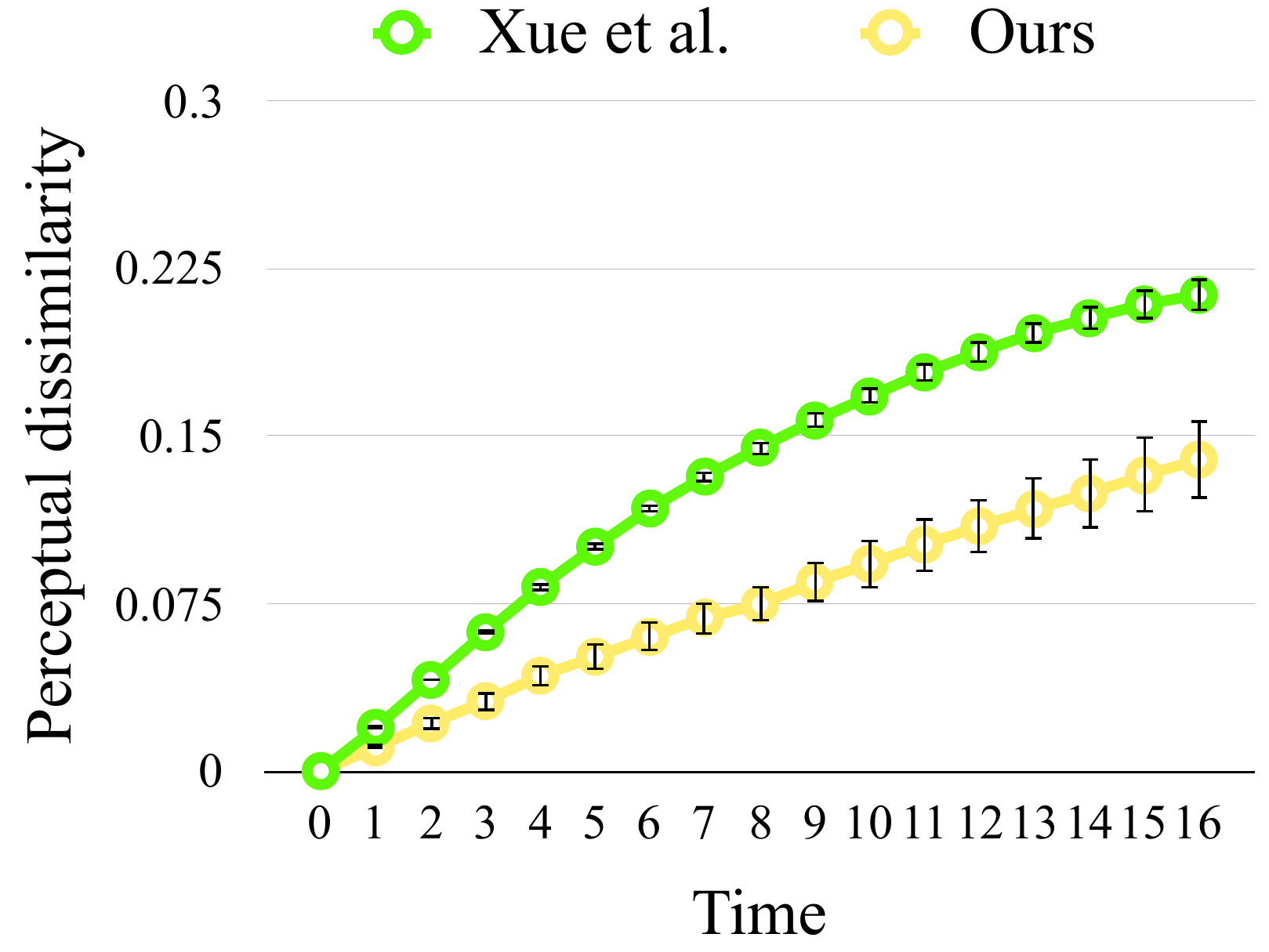} & 
\includegraphics[width = .48\linewidth]{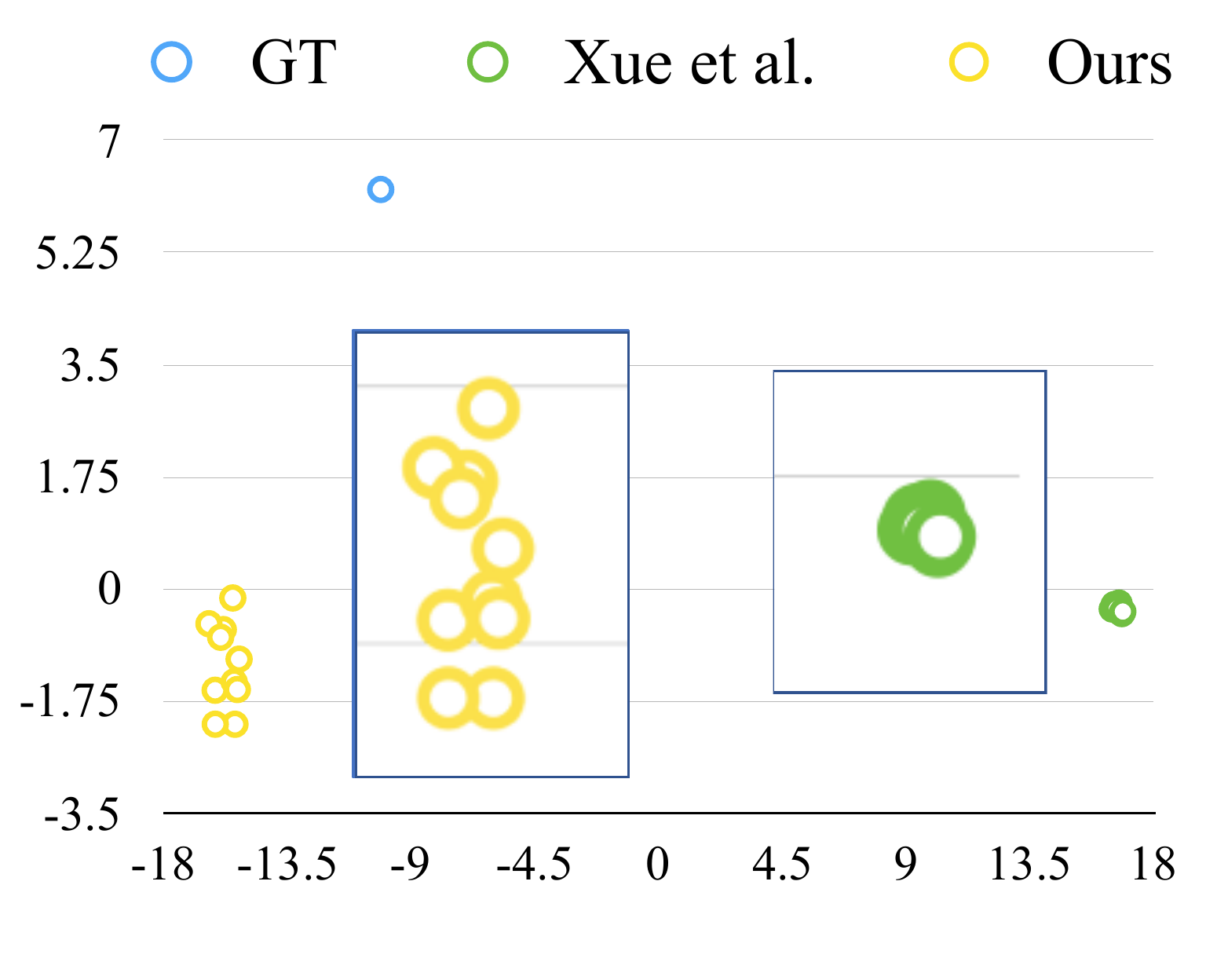} & \\
{(a) Sensitivity on the perceptual quality} & {(b) Visualized distribution } \\
{under different noise} & { of predictions under different noise} \\

\end{tabular}
\caption{Comparisons between~\cite{crossconv-NIPS-2016} and the proposed algorithm on uncertainty modeling given the same starting frame. By drawing different samples, the generated predictions by our method exhibits more diversities while still being more similar to GT. 
}
\label{fig:uncertainty}
\end{figure}

\begin{figure}[t]
\centering
\begin{tabular}{c@{\hspace{0.005\linewidth}}c@{\hspace{0.005\linewidth}}c@{\hspace{0.005\linewidth}}c@{\hspace{0.005\linewidth}}c@{\hspace{0.005\linewidth}}c@{\hspace{0.005\linewidth}}c@{\hspace{0.005\linewidth}}c@{\hspace{0.005\linewidth}}c@{\hspace{0.005\linewidth}}c}

    \includegraphics[width = .103\linewidth]{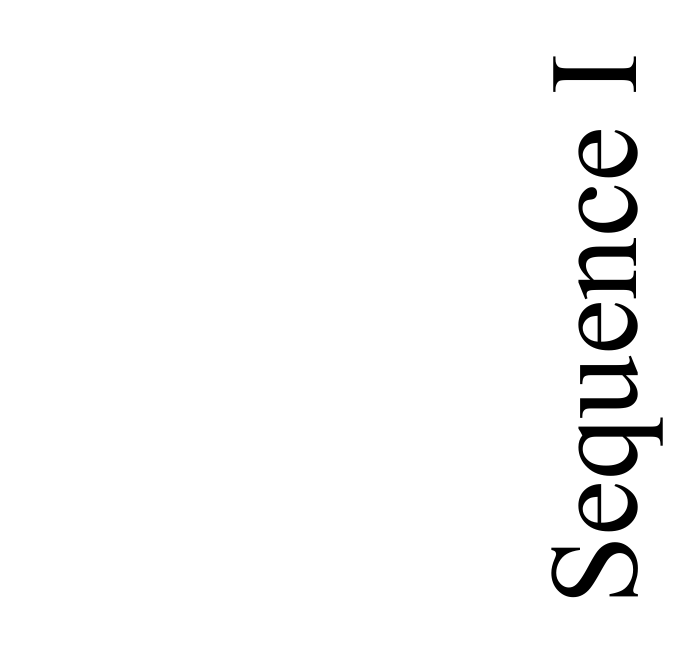} &
    
    \hspace{1pt}\vrule\hspace{1pt}
    
    \includegraphics[width = .103\linewidth]{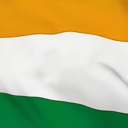} & 
    \includegraphics[width = .103\linewidth]{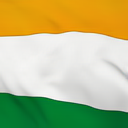} &
    \includegraphics[width = .103\linewidth]{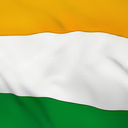} &
    \includegraphics[width = .103\linewidth]{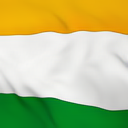} &
    \includegraphics[width = .103\linewidth]{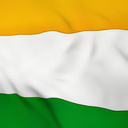} &
    \includegraphics[width = .103\linewidth]{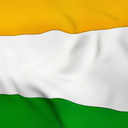} &
    \includegraphics[width = .103\linewidth]{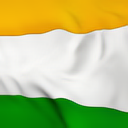} &
    \includegraphics[width = .103\linewidth]{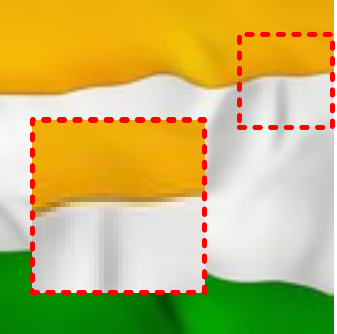} &\\
    
    \includegraphics[width = .103\linewidth]{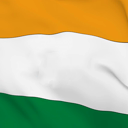} & 
    
    \hspace{1pt}\vrule\hspace{1pt}
    
    \includegraphics[width = .103\linewidth]{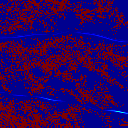} & 
    \includegraphics[width = .103\linewidth]{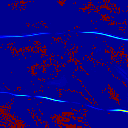} & 
    \includegraphics[width = .103\linewidth]{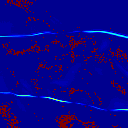} & 
    \includegraphics[width = .103\linewidth]{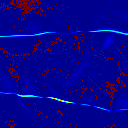} & 
    \includegraphics[width = .103\linewidth]{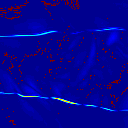} & 
    \includegraphics[width = .103\linewidth]{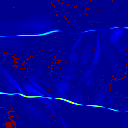} & 
    \includegraphics[width = .103\linewidth]{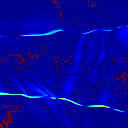} & 
    \includegraphics[width = .103\linewidth]{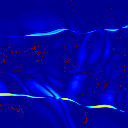} & \\
    
    \includegraphics[width = .103\linewidth]{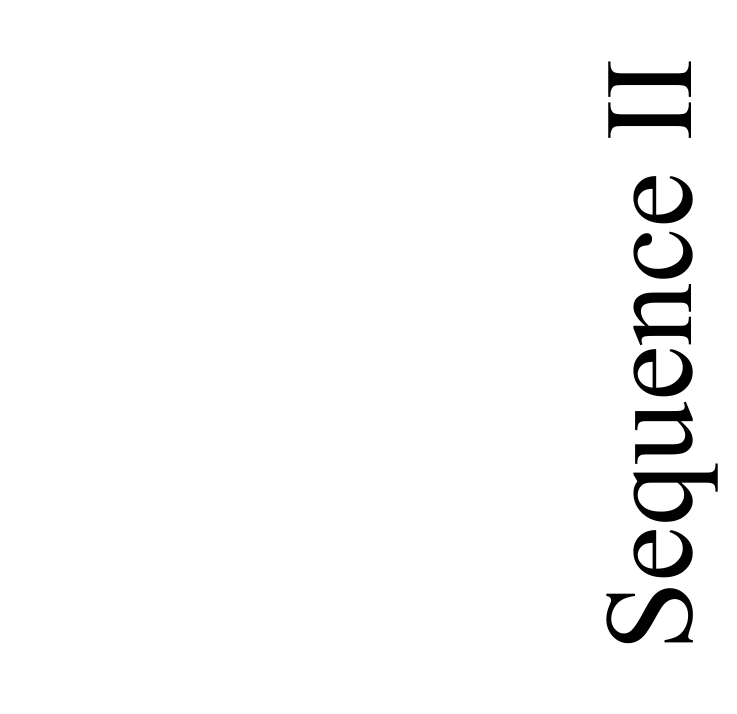} &
    
    \hspace{1pt}\vrule\hspace{1pt}
    
    \includegraphics[width = .103\linewidth]{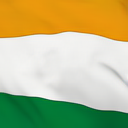} & 
    \includegraphics[width = .103\linewidth]{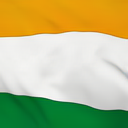} &
    \includegraphics[width = .103\linewidth]{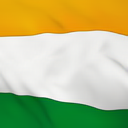} &
    \includegraphics[width = .103\linewidth]{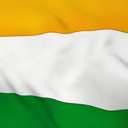} &
    \includegraphics[width = .103\linewidth]{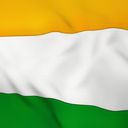} &
    \includegraphics[width = .103\linewidth]{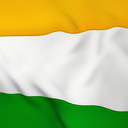} &
    \includegraphics[width = .103\linewidth]{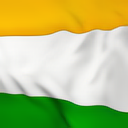} &
    \includegraphics[width = .103\linewidth]{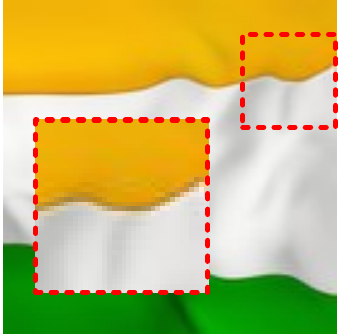} &\\

    { t=0  }& { t=1  }& { t=3  }& { t=5  }& { t=7  }& { t=9  }& { t=11  }& { t=13  }& { t=15  } \\
\end{tabular}
\caption{Given a still image, by sampling different noise in the latent space, our algorithm synthesizes different future outcomes to account for the intrinsic uncertainties. In the middle row, we show the difference of two generated sequences frame-by-frame. 
}
\label{fig:diversity}
\end{figure}

\paragraph{\bf Diversity.}~Both~\cite{crossconv-NIPS-2016} and the proposed method model the uncertainty in predictions, but are different in one-step~\cite{crossconv-NIPS-2016} or multi-step uncertainties. By drawing different samples, we evaluate how the quality of predictions is affected by the noise input and how diverse the predicted sequences are. 
While~\cite{crossconv-NIPS-2016} uses a noise vector of 3200 dimensions and we use that of 2000 dimensions, the noise inputs of two models are not exactly the same but they are all sampled from $N(0,1)$. We sample 10 noise inputs for each method, while ensuring that the two sets of noise inputs have the similar mean and standard deviation. Then we obtain 10 sequences for each method, and compare them with the GT sequence.
Figure~\ref{fig:uncertainty}(a) shows the mean and standard deviation of the perceptual metric over each method's 10 predictions when compared with the GT frame-by-frame. Under different noise inputs, our method keeps generating better sequences that are more similar to the GT. Meanwhile, the results of our algorithm show larger deviation, which implies that there are more diversities in our predictions. To further verify this, we show the embeddings of generated sequences in Figure~\ref{fig:uncertainty}(b). For each sequence, we extract the VGG-19~\cite{VGG-ICLR-2015} features (e.g., fc6 layer) of each frame, stack them as one vector, and map it to a 2-D point through t-SNE~\cite{tsne}.
Figure~\ref{fig:uncertainty}(b) shows that our 10 predictions are much closer to the GT sequence while being scattered to be different from each other. In contrast, the 10 predictions of~\cite{crossconv-NIPS-2016} huddle together and are far from the GT. Those comparisons demonstrate that the proposed algorithm generates more realistic and diverse future predictions. Figure~\ref{fig:diversity} shows an example of two predicted sequences.



\begin{figure}[t]
\centering
\begin{tabular}{c@{\hspace{0.005\linewidth}}c@{\hspace{0.005\linewidth}}c@{\hspace{0.005\linewidth}}c@{\hspace{0.005\linewidth}}c@{\hspace{0.005\linewidth}}c@{\hspace{0.005\linewidth}}c@{\hspace{0.005\linewidth}}c@{\hspace{0.005\linewidth}}c@{\hspace{0.005\linewidth}}c}

    \includegraphics[width = .103\linewidth]{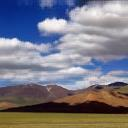} &
    
    \hspace{1pt}\vrule\hspace{1pt}
    
    \includegraphics[width = .103\linewidth]{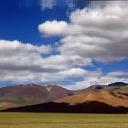} &
    \includegraphics[width = .103\linewidth]{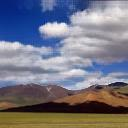} &
    \includegraphics[width = .103\linewidth]{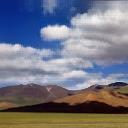} &
    \includegraphics[width = .103\linewidth]{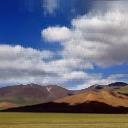} &
    \includegraphics[width = .103\linewidth]{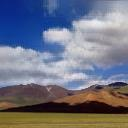} &
    \includegraphics[width = .103\linewidth]{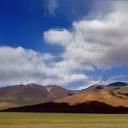} &
    \includegraphics[width = .103\linewidth]{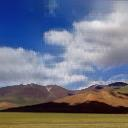} &
    \includegraphics[width = .103\linewidth]{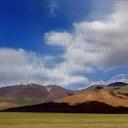} &
    \\
    {t=0}&{t=1}&{t=2}&{t=3}&{t=4}&{t=5}&{t=6}&{t=7}&{t=8} \\

\end{tabular}
\caption{
Potential application of our algorithm in video editing.
}
\label{fig:image2life}
\end{figure}

\paragraph{\bf Bringing still images to life.}~Unlike previous video prediction methods~\cite{mcnet-ICLR-2017,ruben-ICML-2017,walker2016uncertain} that mainly focus on humans for action recognition, our algorithm is more general towards bringing elements in the still image to life, i.e., turning a still image into a vivid GIF for aesthetic effects. It can be an effective tool for video editing. 

In Figure~\ref{fig:image2life}(a), we show a example of turning a photo into a vivid sequence. We mask out the sky region, apply our model trained on the \emph{FloatingCloud} dataset and generate the effect of clouds floating in the sky. This could further benefit existing sky editing methods~\cite{tsai2016sky}.
Moreover, if we replace our flow prediction with known flows from a reference sequence, our flow-to-frame model \emph{Flow2rgb} becomes a global motion style transfer model.
%
%
As the current random sampling strategy for flow predictions is uncontrollable, future work may include introducing more interactions from users to control detailed motions.

\section{Conclusions}

In this work, we propose a video prediction algorithm that synthesizes a set of likely future frames in multiple time steps from one single still image. Instead of directly estimating the high-dimensional future frame space, we choose to decompose this task into a flow prediction phase and a flow-grounded frame generation phase. The flow prediction models the future uncertainty and spatial-temporal relationship in a 3D-cVAE model. The frame generation step helps prevent the manifold shape of predicted sequences from straying off the manifold of real sequences. We demonstrate the effectiveness of the proposed algorithm on both human action videos and dynamic texture videos.

\paragraph{\bf Acknowledgement.}~This work is supported in part by the NSF CAREER Grant \#1149783, gifts from Adobe and NVIDIA. YJL is supported by Adobe and Snap Inc. Research Fellowship.

\clearpage

\bibliographystyle{splncs}
\bibliography{reference}

\clearpage

\section*{Appendix}

\subsection*{Network Architecture}

As shown in the Figure~\ref{fig:framework}(left), our 3D-cVAE model for flow predictions consists of two components, i.e., a 3D variational autoencoder (purple blocks) and an image encoder (pink blocks).
Table~\ref{table:cVAE} shows the detailed architecture of the variational autoencoder (VConv = VolumetricConvolution, VFConv = VolumetricFullConvolution, VBN = VolumetricBatchNormalization, VMP = VolumetricMaxPooling, FN = Filter number, FS = Filter size, S=Stride, P = Padding, \{time, width, height\}).
The \emph{Sampler} is to draw a sample from the latent embedding, same as in~\cite{vae-ICLR-2014,cvae-sohn2015learning}.
The \emph{Mul\_Add} is the conditioning
strategy.

\begin{table}[!htbp]
\vspace{-2mm}
  \caption{Architecture of the variational autoencoder.}
  \label{table:cVAE}
  \centering
  \begin{tabular}{ll}
    \toprule

 Layer ~~ & \\
 \midrule
    $VConv1$~~ & VConv(FN64, FS\{3,3,3\}, S\{1,1,1\}, P\{1,1,1\}), VBN, ReLU \\
     & VMP\{1,2,2\}  \\
    $VConv2$~~ & VConv(FN64, FS\{3,3,3\}, S\{1,1,1\}, P\{1,1,1\}), VBN, ReLU \\
     & VMP\{1,2,2\}  \\
    $VConv3$~~ & VConv(FN128, FS\{3,3,3\}, S\{1,1,1\}, P\{1,1,1\}), VBN, ReLU \\
     & VMP\{2,2,2\}  \\
    $VConv4$~~ & VConv(FN256, FS\{3,3,3\}, S\{1,1,1\}, P\{1,1,1\}), VBN, ReLU \\
     & VMP\{2,2,2\}  \\
    $VConv5$~~ & VConv(FN512, FS\{3,3,3\}, S\{1,1,1\}, P\{1,1,1\}), VBN, ReLU \\
     & VMP\{2,2,2\}  \\
    $MeanVar$~~ & VConv(FN2000, FS\{2,4,4\})  \\
     & VConv(FN2000, FS\{2,4,4\})  \\
    
    $Sampler$~~ &  $\sim$ \\ 
    $Mul\_Add$~~ & $\sim$ \\
    
    $VFConv5$~~ & VFConv(FN512, FS\{2,4,4\}), VBN, ReLU \\
    $VFConv4$~~ & VFConv(FN256, FS\{4,4,4\}, S\{2,2,2\}, P\{1,1,1\}), VBN, ReLU \\
    $VFConv3$~~ & VFConv(FN128, FS\{4,4,4\}, S\{2,2,2\}, P\{1,1,1\}), VBN, ReLU \\
    $VFConv2$~~ & VFConv(FN64, FS\{4,4,4\}, S\{2,2,2\}, P\{1,1,1\}), VBN, ReLU \\
    $VFConv1$~~ & VFConv(FN64, FS\{3,4,4\}, S\{1,2,2\}, P\{1,1,1\}), VBN, ReLU \\
    $Output$~~ & VFConv(FN2, FS\{3,4,4\}, S\{1,2,2\}, P\{1,1,1\})\\
    \bottomrule
  \end{tabular}
\end{table}

Table~\ref{table:ImEncoder} shows the architecture of the image encoder (Conv = SpatialConvolution, \{width, height\}).

\begin{table}[t]
\vspace{-2mm}
  \caption{Architecture of the image encoder.}
  \label{table:ImEncoder}
  \centering
  \begin{tabular}{ll}
    \toprule

 Layer ~~ &  \\
 \midrule
    $Conv1$~~ & Conv(FN64, FS\{4,4\}, S\{2,2\}, P\{1,1\}), ReLU \\
    $Conv2$~~ & Conv(FN64, FS\{4,4\}, S\{2,2\}, P\{1,1\}), ReLU \\
    $Conv3$~~ & Conv(FN128, FS\{4,4\}, S\{2,2\}, P\{1,1\}), ReLU \\
    $Conv4$~~ & Conv(FN256, FS\{4,4\}, S\{2,2\}, P\{1,1\}), ReLU \\
    $Conv5$~~ & Conv(FN512, FS\{4,4\}, S\{2,2\}, P\{1,1\}), ReLU \\
    $Conv6$~~ & Conv(FN2000, FS\{4,4\}) \\
    
    \bottomrule
  \end{tabular}
\end{table}

For the \emph{Flow2rgb} model (Figure~\ref{fig:framework}(right)), the frame and flow encoders (blue and green blocks) share the same architecture as the VGG-19~\cite{VGG-ICLR-2015} up to the Relu\_4\_1 layer, and the decoder (yellow blocks) is designed to be symmetrical to the encoder with the nearest neighbor upsampling layer used for enlarging feature maps.

\end{document}